\documentclass{amsart}
\makeatletter
\tagsleft@false
\makeatother
\usepackage[T1]{fontenc}
\usepackage[utf8]{inputenc}
\usepackage[english]{babel}
\usepackage{times}
\usepackage{geometry}
\usepackage{afterpage}
\usepackage{booktabs}
\usepackage{tabularx}
\usepackage{graphicx}
\usepackage{sidecap}
\usepackage{wrapfig}
\usepackage{subfigure}
\usepackage{setspace}
\usepackage{multicol}
\usepackage{multirow}
\usepackage{verbatim}
\usepackage{xcolor}
\usepackage{textcomp}
\usepackage{cite}
\usepackage{algorithm}
\usepackage{algorithmicx}
\usepackage{algpseudocode}
\usepackage{float}
\usepackage{dirtytalk}
\usepackage{cancel}
\usepackage{chngpage}
\usepackage[hidelinks]{hyperref}
\usepackage{cleveref}

\usepackage{xurl}
\algdef{SE}[DOWHILE]{Do}{doWhile}{\algorithmicdo}[1]{\algorithmicwhile\ #1}

\usepackage{amsmath}
\usepackage{amssymb}
\usepackage{mathtools}
\mathtoolsset{showonlyrefs}
\newcommand{\numberset}{\mathbb}
\newcommand{\N}{\numberset{N}}
\newcommand{\R}{\numberset{R}}

\newcommand{\E}{\numberset{E}}

\renewcommand{\epsilon}{\varepsilon}
\DeclarePairedDelimiter{\abs}{\lvert}{\rvert}
\DeclarePairedDelimiter{\norm}{\lVert}{\rVert}

\newtheorem{thm}{Theorem}

\newtheorem{cor}[thm]{Corollary}

\title{Enhanced uncertainty quantification variational autoencoders for the solution of Bayesian inverse problems}

\author{Andrea Tonini$^\mathrm{1,*}$ \and Luca Dede'$^\mathrm{1,+}$}
\thanks{$^\mathrm{1}$MOX, Dipartimento di Matematica, Politecnico di Milano, Milan, Italy.}
\thanks{$^\mathrm{*}$Corresponding author. E-mail: andrea.tonini@polimi.it}
\thanks{$^\mathrm{+}$E-mail: luca.dede@polimi.it}

\begin{document}
\begin{abstract}
Among other uses, neural networks are a powerful tool for solving deterministic and Bayesian inverse problems in real-time, where variational autoencoders, a specialized type of neural network, enable the Bayesian estimation of model parameters and their distribution from observational data allowing real-time inverse uncertainty quantification. In this work, we build upon existing research [Goh, H. et al., Proceedings of Machine Learning Research, 2022] by proposing a novel loss function to train variational autoencoders for Bayesian inverse problems. When the forward map is affine, we provide a theoretical proof of the convergence of the latent states of variational autoencoders to the posterior distribution of the model parameters. We validate this theoretical result through numerical tests and we compare the proposed variational autoencoder with the existing one in the literature both in terms of accuracy and generalization properties. Finally, we test the proposed variational autoencoder on a Laplace equation, with comparison to the original one and Markov Chains Monte Carlo.
\end{abstract}
\maketitle
\renewcommand{\thefootnote}{}
\footnote{\textbf{Keywords:} Bayesian inverse problem, variational autoencoder, deep learning, partial differential equations.}
\footnote{\textbf{MSC codes:} 68T07, 62C10, 35R30}
\section{Introduction}
Neural networks (NNs) are emerging as an effective tool to approximate the solution of a mathematical model described by means of partial differential equations (PDE)\cite{karniadakis2021physics,du2021evolutional, kast2024positional,regazzoni2024learning,goswami2022physics,lu2021learning}. Observational data can usually be computed as a postprocessing of the PDE solution and can be matched to real data by calibrating the model parameters (e.g., the diffusion coefficient) through the solution of inverse problems. Inverse problems are typically addressed by minimizing a loss function \cite{chong2013introduction,byrd1995limited,nelder1965simplex} to estimate the PDE parameters that best approximate known data. The solution of inverse problems is (much) more computationally demanding than solving the PDE itself, as it requires several model simulations. Moreover, inverse problems are often ill-posed, especially in presence of noisy data, due to the limited amount of data and the large number of model parameters. To address this issue, the loss function can be regularized\cite{tikhonov1977solutions}. Solving inverse problems by using neural networks as a surrogate model has garnered significant interest in several fields (e.g., applications in life sciences and computational medicine \cite{salvador2024digital,salvador2024whole,caforio2024physics}) as neural networks allow for fast model simulation, indirectly reducing the computational cost of minimizing the loss function.

Inverse problems can be solved more efficiently by using neural networks as surrogate solvers rather than surrogate models. Tikhonov Networks (TNets) have been shown to effectively solve deterministic inverse problems \cite{nguyen2024tnet}. The loss function of TNets incorporates a priori parameter knowledge and a physics-aware regularization term. Furthermore, data randomization has been both theoretically and numerically shown to indirectly enforce additional regularization of the neural network's weights and biases. Moreover, a new Tikhonov autoencoder neural network (TAEN) framework has improved the performances of TNets \cite{nguyen5081218taen}. A NETwork Tikhonov (NETT) has been applied to inverse problems involving images, using a data-driven regularization approach \cite{li2020nett}. Convolutional Neural Networks (CNNs) have also been extensively studied in the context of image-based inverse problems \cite{jin2017deep,adler2017solving,lucas2018using}.

In the statistical framework, Bayesian inverse problems account for noisy data, reflecting realistic scenarios. The objective is to estimate both the parameters of interest and their associated uncertainties, which makes Bayesian inverse problems more computationally expensive than their deterministic counterpart. Variational AutoEncoders (VAE) have been used to explicitly generate the prior distribution of the parameters \cite{gonzalez2019solving}. In \cite{jiang2022combining}, an encoder is used to reduce the dimensionality of the parameter space dimension, after which a Markov Chain Monte Carlo method is applied to solve a Bayesian inverse problem in the reduced space. The full parameter space is subsequently recovered using a decoder. In \cite{hou2019solving}, a class of network training methods combined with sample-based Bayesian inverse algorithms is proposed. Conditional VAEs have been applied to perform variational inference of the posterior parameter distribution \cite{gabbard2022bayesian,tonolini2020variational}, where the loss function represents an upper bound of the cross-entropy between the posterior distribution and its approximation. In \cite{chua2020learning}, neural networks integrated with reduced order models are used to solve Bayesian inverse problems on a reduced parameter space. Uncertainty Quantification Variational AutoEncoders (UQ-VAE) have been shown to effectively address Bayesian inverse problems by estimating the mean and covariance of the posterior distribution \cite{goh2021solving}. This approach leverages on a family of Jensen-Shannon divergences \cite{nielsen2010family} to construct a loss function, but it lacks an accurate theoretical result. Specifically, the proven theoretical result applies only to a lower bound of a quantity which involves expected values. These expectations are then approximated using a single Monte Carlo sample to define the UQ-VAE loss function, thereby weakening the connection between the theoretical result and the actual training loss. The Jensen-Shannon divergence is extensively used in generative modeling \cite{deasy2020constraining,sutter2020multimodal}, where the posterior distribution of parameters is used to generate new samples.

This work proposes an enhanced Uncertainty Quantification Variational AutoEncoder (eUQ-VAE) to improve the parameter posterior distribution estimate of UQ-VAEs. We define a new loss function for the variational autoencoder and, when the forward problem is affine, prove the convergence of transformed latent states of the eUQ-VAE to the mean and covariance of the posterior distribution. Numerical tests validate this theoretical result, showcase the generalization capabilities of eUQ-VAEs and their performances on a Laplace problem.

The paper is organized as follows: in \Cref{sec:buildingLossFunction}, we introduce the eUQ-VAE approach and derive the training loss function, together with theoretical results for affine forward problems that guarantee the convergence of a transformation of the latent states to an estimate of the model parameters and their uncertainty; in \Cref{sec:results} we present numerical tests that validate the theoretical results, compare the eUQ-VAE approach with the UQ-VAE one and show the performance of eUQ-VAE on a non-affine Laplace Bayesian inverse problem; finally, \Cref{sec:concl} summarizes the conclusions of this work.

\section{Bayesian inverse problems} \label{sec:buildingLossFunction}
We describe the general framework of Bayesian inverse problems and introduce a new method for solving them using variational autoencoders.

Let $U$ and $Y$ be random variables representing a set of parameters and a set of observational data, respectively. We assume that there exists a function $\mathcal F$ that maps the parameters to the observational data. Additionally, we assume that the observational data are subject to random noise $E$. The mathematical model under consideration is:
\begin{align}
Y = \mathcal F(U)+E. \label{eq:model}
\end{align}

The solution to a Bayesian inverse problem is the probability distribution of the parameters $\mathbf u \in \R^\mathrm{D}$ conditioned on a set of observational data $\mathbf y \in \R^\mathrm{O}$, where $D,O \in \N$. This is known as the posterior distribution $p_\mathrm{U|Y}(\mathbf u | \mathbf y)$. Note that $\mathcal F: \R^\mathrm{D} \to \R^\mathrm{O} $. Using the Bayes' theorem \cite{bayes1958essay}, the posterior distribution is typically maximized with respect to the parameters $\mathbf u$, yielding the Maximum A Posteriori estimate (MAP) $\mathbf u_\mathrm{MAP} \in \R^\mathrm{D}$, which represents the most likely parameter to generate the given $\mathbf y$. By Bayes' theorem:
\begin{align}
p_\mathrm{U|Y}(\mathbf u | \mathbf y) = \frac{p_\mathrm{U,Y}(\mathbf u, \mathbf y)}{p_\mathrm{Y}(\mathbf y)} = \frac{p_\mathrm{Y|U}(\mathbf y | \mathbf u)\,p_\mathrm{U}(\mathbf u)}{p_\mathrm{Y}(\mathbf y)}, \label{eq:postProb}
\end{align} 
where $p_\mathrm{U,Y}(\mathbf u, \mathbf y)$, $p_\mathrm{Y}(\mathbf y)$, $p_\mathrm{U}(\mathbf u)$ and $p_\mathrm{Y|U}(\mathbf y | \mathbf u)$ represent the joint probability density function (pdf) of $U$ and $Y$, the marginal pdf of $Y$, the marginal pdf of $U$ and the conditional pdf of $Y$ given $U$ (likelihood), respectively. Assuming that the random noise $E$ is independent of $U$, the likelihood can be rewritten as:
\begin{align}
p_\mathrm{Y|U}(\mathbf y | \mathbf u) = p_\mathrm{E}(\mathbf y-\mathcal F(\mathbf u)), \label{eq:noiseProb}
\end{align}
where $p_\mathrm{E}$ is the pdf of the random variable E. Furthermore, we assume that $U \sim \mathcal N(\boldsymbol \mu_\mathrm{pr}, \Gamma_\mathrm{pr})$ and $E \sim \mathcal N(\boldsymbol \mu_\mathrm{E}, \Gamma_\mathrm{E})$, where $\boldsymbol \mu_\mathrm{pr} \in \R^{D}$ and $\boldsymbol \mu_\mathrm{E} \in \R^{O}$ represent the means of the prior and error distributions, respectively, and $\Gamma_\mathrm{pr} \in \R^\mathrm{D \times D}$ and $\Gamma_\mathrm{E} \in \R^\mathrm{O \times O}$ are the corresponding covariance matrices. In this work, we consider only symmetric positive definite (SPD) covariance matrices, as well as their inverses. Using \eqref{eq:noiseProb}, the joint pdf of $U$ and $Y$can be rewritten as:
\begin{multline}
 p_\mathrm{U,Y}(\mathbf u, \mathbf y) = p_\mathrm{Y|U}(\mathbf y | \mathbf u)p_\mathrm{U}(\mathbf u)  =  \frac{1}{(2\pi)^\mathrm{\frac{O+D}{2}}|\Gamma_\mathrm{E}|^\mathrm{\frac{1}{2}} |\Gamma_\mathrm{pr}|^\mathrm{\frac{1}{2}}}\\
\text{exp}\left( -\frac{1}{2}\left( \norm{\mathbf y-\mathcal F(\mathbf u) -\boldsymbol \mu_\mathrm{E}}^\mathrm{2}_\mathrm{\Gamma_\mathrm{E}^\mathrm{-1}} +  \norm{\mathbf u -\boldsymbol \mu_\mathrm{pr}}^\mathrm{2}_\mathrm{\Gamma_\mathrm{pr}^\mathrm{-1}} \right) \right),
\end{multline}
where $| \cdot |$ denotes the determinant of a matrix  and $\norm{\cdot}_\mathrm{\Gamma^\mathrm{-1}}$ is the norm induced by a SPD matrix $\Gamma^\mathrm{-1}$. The norms appearing in the exponent are adimensionalized thanks to the matrix weighted norm. Maximizing the posterior pdf $p_\mathrm{U|Y}(\mathbf u | \mathbf y)$, we obtain the MAP estimate:
\begin{align}
\mathbf u_\mathrm{MAP} = \underset{\mathbf u \in \R^\mathrm{D}}{\mathrm{argmax}}\, p_\mathrm{U|Y}(\mathbf u | \mathbf y) = \underset{\mathbf u \in \R^\mathrm{D}}{\mathrm{argmin}}\left( \norm{\mathbf y-\mathcal F(\mathbf u) -\boldsymbol \mu_\mathrm{E}}^\mathrm{2}_\mathrm{\Gamma_\mathrm{E}^\mathrm{-1}} +  \norm{\mathbf u -\boldsymbol \mu_\mathrm{pr}}^\mathrm{2}_\mathrm{\Gamma_\mathrm{pr}^\mathrm{-1}}\right).
\label{eq:mapMu}
\end{align}
The first term of the optimization problem, given by the likelihood model, represents the mismatch between observed data and the parameters to output map evaluations, while the second term incorporates prior knowledge about the parameters and acts as a regularization term. Solving \eqref{eq:mapMu} typically requires an iterative gradient-based method, which can be computationally expensive. To perform inverse uncertainty quantification (UQ), the covariance matrix of the posterior distribution is often approximated using the Laplace approximation \cite{evans2000approximating}:
\begin{align}
\Gamma_\mathrm{Lap} = \left(J_\mathrm{\mathcal F}(\mathbf u_\mathrm{MAP})^\mathrm{T} \Gamma_\mathrm{E}^\mathrm{-1}J_\mathrm{\mathcal F}(\mathbf u_\mathrm{MAP}) + \Gamma_\mathrm{pr}^\mathrm{-1} \right)^\mathrm{-1}, \label{eq:mapGamma}
\end{align}
where $J_\mathrm{\mathcal F}(\mathbf u_\mathrm{MAP}) \in \R^\mathrm{O \times D}$ is the Jacobian of $\mathcal F$ evaluated at $\mathbf u_\mathrm{MAP}$. Thus, computing $\Gamma_\mathrm{Lap}$ requires evaluating $J_\mathrm{\mathcal F}(\mathbf u_\mathrm{MAP})$, which is generally expensive.

The high computational cost of solving Bayesian inverse problems motivates the need for a more efficient methodology. We aim to develop an alternative optimization problem to find both $(\mathbf u_\mathrm{MAP}, \Gamma_\mathrm{Lap})$. Given the observational data $\mathbf y$, we seek to approximate $p_\mathrm{U|Y}(\mathbf u | \mathbf y)$ using a Gaussian distribution $q_\mathrm{\phi}(\mathbf u | \mathbf y) = \mathcal N (\boldsymbol \mu_\mathrm{post}(\mathbf y,\phi),\Gamma_\mathrm{post}(\mathbf y,\phi))$, where $\boldsymbol \mu_\mathrm{post} \in \R^\mathrm{D}$ and $\Gamma_\mathrm{post}\in \R^\mathrm{D \times D}$ are the mean and the covariance matrix that depend on $\mathbf y$ and $\phi$, a set of hyper-parameters. For brevity, we omit to explicit the dependence of $(\boldsymbol \mu_\mathrm{post},\Gamma_\mathrm{post})$ on $\mathbf y$ and $\phi$. To solve a Bayesian inverse problem we build a new loss function adapting the procedure proposed in \cite{goh2021solving}:
\begin{enumerate}
\item Introduce a proxy distribution $q_\mathrm{pro,\phi}(\mathbf u| \mathbf y) = \mathcal N(\boldsymbol \mu(\mathbf y,\phi), \Gamma(\mathbf y,\phi))$ to obtain $q_\mathrm{\phi}(\mathbf u| \mathbf y)$, with mean $\boldsymbol \mu \in \R^\mathrm{D}$ and covariance matrix $\Gamma \in \R^\mathrm{D \times D}$, that depend on $\mathbf y$ and $\phi$, a set of hyper-parameters.
\item Measure the difference between the target posterior distribution $p_\mathrm{U|Y}(\mathbf u | \mathbf y)$ and the proxy distribution $q_\mathrm{pro,\phi}(\mathbf u | \mathbf y)$ by means of a family of Jensen-Shannon divergences (JSD) \cite{nielsen2010family}. This difference is the starting point to build the loss function to minimize (\Cref{sec:JSD}).
\item Build an upper bound of the JSD and an additional Kullback-Leibler divergence (KLD) \cite{kullback1951information} term, which will serve as the loss function for the optimization problem (\Cref{sec:upperBound}).
\item  When $\mathcal F$ is affine, show that the stationary points of the upper bound with respect to $(\boldsymbol \mu,\Gamma)$ can be used to compute $(\mathbf u_\mathrm{MAP}, \Gamma_\mathrm{Lap})$. This step establishes theoretical foundations for our method and to determine how to compute $(\mathbf u_\mathrm{MAP}, \Gamma_\mathrm{Lap})$ (\Cref{sec:anRes}).
\item Solve the Bayesian inverse problem using an uncertainty quantification variational autoencoder (UQ-VAE) \cite{goh2021solving}, a type of neural network. In this case, $\phi$ represents the weights and biases of the UQ-VAE (\Cref{sec:UQVAE}).
\end{enumerate}  

\subsection{Jensen-Shannon divergences} \label{sec:JSD}
We aim to redefine the optimization problem in \eqref{eq:mapMu} because the computation of $\Gamma_\mathrm{Lap}$ involves evaluating $J_\mathrm{\mathcal F}(\mathbf u_\mathrm{MAP})$.

Consider two probability spaces $(\R^\mathrm{D},\mathcal B(\R^\mathrm{D}),\nu_\mathrm{i})$ $i=1,2$, where $\mathcal B(\R^\mathrm{D})$ is the Borel $\sigma$-algebra and $\nu_\mathrm{1},\nu_\mathrm{2}$ are two probability measures. Assume that $\nu_1$ and $\nu_2$ are absolutely continuous with respect to each other ($\nu_1 \equiv \nu_2$) and let $\lambda \equiv \nu_\mathrm{1}$ and $\lambda \equiv \nu_\mathrm{2}$. According to the Radon-Nikodym theorem, there exist two unique (up to sets of null measure) pdfs, $q(\mathbf u)$ and $p(\mathbf u)$ associated with $\nu_\mathrm{1}$ and $\nu_\mathrm{2}$, respectively, which are $\infty > q(\mathbf u),p(\mathbf u)>0$ $\lambda$-almost surely. The KLDs are then well-defined as:
\begin{align}
KL(q||p) = \E_\mathrm{q}\left[\log\left( \frac{q(\mathbf u)}{p(\mathbf u)}\right) \right], \qquad
KL(p||q) = \E_\mathrm{p}\left[\log\left( \frac{p(\mathbf u)}{q(\mathbf u)}\right) \right] ,
\end{align}
where $\E_\mathrm{q}$ denotes the expected value with respect to the probability measure $\nu_\mathrm{1}$ associated with $q$. It is a known property that $KL(q||p) \ge 0$, with equality holding if and only if $q=p$ $\nu_\mathrm{1}$-almost surely (since $\nu_\mathrm{1} \equiv \nu_\mathrm{2}$ the equality holds also $\nu_\mathrm{2}$-almost surely). We consider a family of JSDs defined for $\alpha \in [0,1]$:
\begin{align}
JS_\mathrm{\alpha}(q||p) = \alpha KL(q || (1-\alpha) q + \alpha p) + (1-\alpha) KL(p || (1-\alpha) q + \alpha p).
\label{eq:JSDfamily}
\end{align}
By varying $\alpha$, this formula allows us to explore a range of JSDs. Note that the second argument of each KLD is a convex combination of $q$ and $p$, ensuring that the KLDs remain well-defined. Additionally, the convex combination of the two KLDs ensures that $JS_\mathrm{\alpha}(q||p)\ge 0$. It is straightforward to verify that $JS_\mathrm{\alpha}(q||p)= 0$ if and only if $q = p$ $\nu_\mathrm{1}$- and $\nu_\mathrm{2}$-almost surely.

We state a theorem from \cite{goh2021solving} that is relevant to our discussion.
\begin{thm}
Let $\alpha \in [0,1]$. Then
\begin{align}
JS_\mathrm{\alpha}(q_\mathrm{pro,\phi}(\mathbf u | \mathbf y) ||p_\mathrm{U|Y} (\mathbf u | \mathbf y) ) = &-\alpha \E_\mathrm{ q_\mathrm{pro,\phi}(\mathbf u | \mathbf y)}\left[ \log\left( \alpha + \frac{(1-\alpha)q_\mathrm{pro,\phi}(\mathbf u | \mathbf y)}{p_\mathrm{U|Y}(\mathbf u| \mathbf y)}\right)  \right] \label{eq:JSDth}\\
 &+ \alpha \log(p_\mathrm{Y}(\mathbf y)) \\
 &- (1-\alpha) \E_\mathrm{p_\mathrm{U|Y}(\mathbf u| \mathbf y)} \left[  \log\left( \alpha + \frac{(1-\alpha)q_\mathrm{pro,\phi}(\mathbf u | \mathbf y)}{p_\mathrm{U|Y}(\mathbf u| \mathbf y)}\right) \right]\\
 & -\alpha \E_\mathrm{q_\mathrm{pro,\phi}(\mathbf u | \mathbf y)} \left[ \log \left( \frac{p_\mathrm{U,Y}(\mathbf u, \mathbf y)}{q_\mathrm{pro,\phi}(\mathbf u | \mathbf y)}\right)  \right].
\end{align}
\end{thm}
 
This theorem provides an alternate expression of the family of JSDs. However, in practice, minimizing this expression with respect to $q_\mathrm{pro,\phi}(\mathbf u | \mathbf y)$ is unfeasible because it involves the unknown $p_\mathrm{U|Y}(\mathbf u| \mathbf y)$ in a non-separable manner.

\subsection{Upper bound for the JSD family} \label{sec:upperBound}
In this section, we define a computable function that provides the loss function of the optimization problem we will solve.
\begin{cor}\label{cor:upperBound}
Let $\alpha \in [0,1)$. Then
\begin{align}
JS_\mathrm{\alpha}(q_\mathrm{pro,\phi}(\mathbf u | \mathbf y) ||p_\mathrm{U|Y} (\mathbf u | \mathbf y) )  \le & -\alpha KL(q_\mathrm{pro,\phi}(\mathbf u|\mathbf y) || p_\mathrm{U|Y}(\mathbf u|\mathbf y))\\
&+\alpha \log(p_\mathrm{Y}(\mathbf y)) -\log(1-\alpha)\\
&+(1-\alpha)KL(p_\mathrm{U|Y}(\mathbf u|\mathbf y) || q_\mathrm{pro,\phi}(\mathbf u|\mathbf y))\\
&-\alpha \E_\mathrm{q_\mathrm{pro,\phi}(\mathbf u | \mathbf y)}\left[ \log(p_\mathrm{Y|U}(\mathbf y | \mathbf u)) \right] + \alpha KL(q_\mathrm{pro,\phi}(\mathbf u | \mathbf y) || p_\mathrm{U}(\mathbf u)). \label{eq:cor1term}
\end{align}
\end{cor}
 
The proof is given in \cite{goh2021solving}. Moving the term $\alpha KL(q_\mathrm{pro,\phi}(\mathbf u|\mathbf y) || p_\mathrm{U|Y}(\mathbf u|\mathbf y))$ of \eqref{eq:cor1term} to the left hand side of the inequality, we obtain that:
\begin{gather}
LB_\mathrm{\alpha}(q_\mathrm{pro,\phi} ) \coloneq JS_\mathrm{\alpha}(q_\mathrm{pro,\phi}(\mathbf u | \mathbf y) ||p_\mathrm{U|Y} (\mathbf u | \mathbf y) ) + \alpha KL(q_\mathrm{pro,\phi}(\mathbf u|\mathbf y) || p_\mathrm{U|Y}(\mathbf u|\mathbf y)) \\
 \le \alpha \log(p_\mathrm{Y}(\mathbf y)) -\log(1-\alpha)+(1-\alpha)\E_\mathrm{p_\mathrm{U|Y}(\mathbf u|\mathbf y)}[\log p_\mathrm{U|Y}(\mathbf u|\mathbf y) ]\\
- (1\!-\! \alpha) \E_\mathrm{p_\mathrm{U|Y}(\mathbf u|\mathbf y)}\! \left[ \log (q_\mathrm{pro,\phi}(\mathbf u|\mathbf y)) \right] \!-\! \alpha \E_\mathrm{q_\mathrm{pro,\phi}(\mathbf u | \mathbf y)}\! \left[ \log(p_\mathrm{Y|U}(\mathbf y | \mathbf u)) \right]
\! +\! \alpha KL(q_\mathrm{pro,\phi}(\mathbf u | \mathbf y) || p_\mathrm{U}(\mathbf u)). \label{eq:lowerBound}
\end{gather}
Observe that $LB_\mathrm{\alpha}(q_\mathrm{pro,\phi})$ is still bigger than or equal to $0$ and it is null if and only if $q_\mathrm{pro,\phi}(\mathbf u|\mathbf y) = p_\mathrm{U|Y}(\mathbf u|\mathbf y)$ almost surely as it holds for the JSD family \eqref{eq:JSDfamily}. We will define an upper bound for $LB_\mathrm{\alpha}(q_\mathrm{pro,\phi})$ that will constitute the loss function.

We analyze separately each term of \eqref{eq:lowerBound}, by providing the theoretical details for each inequality involved in the construction of the loss function.\\ 
To estimate $ - \E_\mathrm{p_\mathrm{U|Y}(\mathbf u|\mathbf y)}  \left[ \log (q_\mathrm{pro,\phi}(\mathbf u|\mathbf y)) \right]$ in \eqref{eq:lowerBound}, we use, in order, the normality of $q_\mathrm{pro,\phi}(\mathbf u| \mathbf y)$, the definition of the expected value, Bayes' theorem to obtain \eqref{eq:postProb} and the independence between $U$ and $E$ to get \eqref{eq:noiseProb}, the normality of $E$ and, finally, rewrite the expected value in the last equality by exploiting that $\Gamma$ is SPD:
\begin{gather}
 - \E_\mathrm{p_\mathrm{U|Y}(\mathbf u|\mathbf y)}  \left[ \log (q_\mathrm{pro,\phi}(\mathbf u|\mathbf y)) \right] = \frac{D}{2}\log(2\pi)+\frac{ \log |\Gamma|}{2} +  \frac{1}{2}\E_\mathrm{p_\mathrm{U|Y}(\mathbf u|\mathbf y)} \left[ \norm{\boldsymbol \mu-\mathbf u}_\mathrm{\Gamma^\mathrm{-1}}^\mathrm{2} \right] \\
 = \frac{D}{2}\log(2\pi)+ \frac{1}{2}\left( \log |\Gamma| + \int_\mathrm{\R^\mathrm{D}} \norm{\boldsymbol \mu - \mathbf u}_\mathrm{\Gamma^\mathrm{-1}}^\mathrm{2} p_\mathrm{U|Y}(\mathbf u | \mathbf y) d\mathbf u \right) \\
 =  \frac{D}{2}\log(2\pi)+ \frac{1}{2}\left(\log |\Gamma| + \int_\mathrm{\R^\mathrm{D}} \norm{\boldsymbol \mu-\mathbf u}_\mathrm{\Gamma^\mathrm{-1}}^\mathrm{2} \frac{ p_\mathrm{E}(\mathbf y-\mathcal F(\mathbf u)) p_\mathrm{U}(\mathbf u)}{p_\mathrm{Y}(\mathbf y)} d\mathbf u \right)\\
 = \frac{D}{2}\log(2\pi)+ \frac{1}{2}\left( \log |\Gamma| + \frac{1}{p_\mathrm{Y}(\mathbf y)}\int_\mathrm{\R^\mathrm{D}} \norm{\boldsymbol \mu-\mathbf u}_\mathrm{\Gamma^\mathrm{-1}}^\mathrm{2} 
\frac{e^\mathrm{-\frac{1}{2}\norm{\mathbf y-\mathcal F(\mathbf u) -\boldsymbol \mu_\mathrm{E} }_\mathrm{\Gamma_\mathrm{E}^\mathrm{-1}}^\mathrm{2}}}{(2\pi)^\mathrm{\frac{O}{2}} |\Gamma_\mathrm{E}|^\mathrm{\frac{1}{2}}}
p_\mathrm{U}(\mathbf u) d\mathbf u \right) \\
 \le \frac{D}{2}\log(2\pi)+ \frac{1}{2}\left(\log |\Gamma| +\frac{1}{p_\mathrm{Y}(\mathbf y)(2\pi)^\mathrm{\frac{O}{2}} |\Gamma_\mathrm{E}|^\mathrm{\frac{1}{2}}}\E_\mathrm{p_\mathrm{U}(\mathbf u)}\left[  \norm{\boldsymbol \mu - \mathbf u}_\mathrm{\Gamma^\mathrm{-1}}^\mathrm{2} \right] \right)\\
 = \frac{D}{2}\log(2\pi)+ \frac{1}{2}\left(\log |\Gamma| + \frac{1}{p_\mathrm{Y}(\mathbf y)(2\pi)^\mathrm{\frac{O}{2}} |\Gamma_\mathrm{E}|^\mathrm{\frac{1}{2}}} \left( \norm{\boldsymbol \mu - \boldsymbol \mu_\mathrm{pr}}_\mathrm{\Gamma^\mathrm{-1}}^\mathrm{2} + \mathrm{tr}\left(\Gamma^\mathrm{-1}\Gamma_\mathrm{pr}  \right) \right) \right), \label{eq:pUexpval}
\end{gather}
where $\mathrm{tr}$ is the trace operator.

We rewrite $-\E_\mathrm{q_\mathrm{pro,\phi}(\mathbf u | \mathbf y)}\left[ \log(p_\mathrm{Y|U}(\mathbf y | \mathbf u)) \right]$ in \eqref{eq:lowerBound} by using \eqref{eq:noiseProb} for $U$ and $E$ independent, the normality of $E$, with $\Gamma_\mathrm{E}$ SPD:
\begin{gather}
-\E_\mathrm{q_\mathrm{pro,\phi}(\mathbf u | \mathbf y)}\left[ \log(p_\mathrm{Y|U}(\mathbf y | \mathbf u)) \right]  = \frac{O}{2}\log (2 \pi) + \frac{\log|\Gamma_\mathrm{E}|}{2}+ \frac{1}{2} \E_\mathrm{q_\mathrm{pro,\phi}(\mathbf u | \mathbf y)}\left[ \norm{\mathbf y - \boldsymbol \mu_\mathrm{E} - \mathcal F(\mathbf u)}_\mathrm{\Gamma_\mathrm{E}^\mathrm{-1}}^\mathrm{2} \right]\\
 =  \frac{O}{2}\log (2 \pi) \!+\! \frac{\log|\Gamma_\mathrm{E}|}{2}\!+\! \frac{1}{2}\left(\norm{\mathbf y\! -\! \boldsymbol \mu_\mathrm{E}\! -\! \E_\mathrm{q_\mathrm{pro,\phi}(\mathbf u | \mathbf y)}\! \left[ \mathcal F(\mathbf u) \right]}_\mathrm{\Gamma_\mathrm{E}^\mathrm{-1}}^\mathrm{2}\! +\! \mathrm{tr} \left( \Gamma_\mathrm{E}^\mathrm{-1} \mathrm{Cov}_\mathrm{q_\mathrm{pro,\phi}(\mathbf u | \mathbf y)}\! \left[ \mathcal F(\mathbf u)  \right] \right)\right)\! . \label{eq:qUexpval} 
\end{gather}

For $KL(q_\mathrm{pro,\phi}(\mathbf u | \mathbf y) || p_\mathrm{U}(\mathbf u))$ in \eqref{eq:lowerBound}, by using the normality of $U$ and $q_\mathrm{pro,\phi}(\mathbf u| \mathbf y)$, we get:
\begin{align}
 KL(q_\mathrm{pro,\phi}(\mathbf u | \mathbf y) || p_\mathrm{U}(\mathbf u)) &= \E_\mathrm{q_\mathrm{pro,\phi}(\mathbf u | \mathbf y)}\left[ \log \left(\frac{q_\mathrm{pro,\phi}(\mathbf u | \mathbf y)}{ p_\mathrm{U}(\mathbf u)} \right) \right]\\
 & = \frac{1}{2} \left( \log \frac{|\Gamma_\mathrm{pr}|}{|\Gamma|} +\mathrm{tr}\left( \Gamma_\mathrm{pr}^\mathrm{-1} \Gamma  \right) + \norm{\boldsymbol \mu - \boldsymbol \mu_\mathrm{pr}}_\mathrm{\Gamma_\mathrm{pr}^\mathrm{-1}}^\mathrm{2} -D  \right).
\end{align}

Combining all the terms, an upper bound of \eqref{eq:lowerBound} is given by:
\begin{gather}
C_\mathrm{1} + C_\mathrm{2}\Bigl( (1-\alpha) \left( \log |\Gamma| + \norm{\boldsymbol \mu - \boldsymbol \mu_\mathrm{pr}}_\mathrm{\Gamma^\mathrm{-1}}^\mathrm{2} + \mathrm{tr}\left(\Gamma^\mathrm{-1}\Gamma_\mathrm{pr}  \right) \right)  \\
+\alpha  \E_\mathrm{q_\mathrm{pro,\phi}(\mathbf u | \mathbf y)}\left[ \norm{\mathbf y - \boldsymbol \mu_\mathrm{E} - \mathcal F(\mathbf u)}_\mathrm{\Gamma_\mathrm{E}^\mathrm{-1}}^\mathrm{2} \right] \\
  + \alpha \left(  -\log |\Gamma|  + \norm{\boldsymbol \mu - \boldsymbol \mu_\mathrm{pr}}_\mathrm{\Gamma_\mathrm{pr}^\mathrm{-1}}^\mathrm{2} +\mathrm{tr}\left( \Gamma_\mathrm{pr}^\mathrm{-1} \Gamma  \right)  \right) \Bigr), \label{eq:upperBoundFinal}
\end{gather}
where $C_\mathrm{1}$ and $C_\mathrm{2}$ are constants in $(\boldsymbol \mu,\Gamma)$ given by:
\begin{align}
C_\mathrm{1} = & \alpha \log(p_\mathrm{Y}(\mathbf y)) -\log(1-\alpha)+(1-\alpha)\E_\mathrm{p_\mathrm{U|Y}(\mathbf u|\mathbf y)}[\log p_\mathrm{U|Y}(\mathbf u|\mathbf y) ] \\
 &+ \frac{(1-\alpha)}{2}D\log(2\pi) + \frac{\alpha}{2} \left( O\log(2\pi)+\log \left(|\Gamma_\mathrm{E}| |\Gamma_\mathrm{pr}| \right) -D\right),\\
C_\mathrm{2} = & \frac{1}{2}\max \left \{ 1,\frac{1}{p_\mathrm{Y}(\mathbf y)(2\pi)^\mathrm{\frac{O}{2}} |\Gamma_\mathrm{E}|^\mathrm{\frac{1}{2}}} \right \}.
\end{align}
We remove from the upper bound \eqref{eq:upperBoundFinal} the logarithmic terms as they may cause instabilities during its minimization. We define:
\begin{gather}
 UB_\mathrm{\alpha}(q_\mathrm{pro,\phi} ) \coloneq C_\mathrm{1} + C_\mathrm{2}\Bigl( (1-\alpha) \left( \norm{\boldsymbol \mu - \boldsymbol \mu_\mathrm{pr}}_\mathrm{\Gamma^\mathrm{-1}}^\mathrm{2} + \mathrm{tr}\left(\Gamma^\mathrm{-1}\Gamma_\mathrm{pr}  \right) \right)  \\
+\alpha  \E_\mathrm{q_\mathrm{pro,\phi}(\mathbf u | \mathbf y)}\left[ \norm{\mathbf y - \boldsymbol \mu_\mathrm{E} - \mathcal F(\mathbf u)}_\mathrm{\Gamma_\mathrm{E}^\mathrm{-1}}^\mathrm{2} \right] \\
  + \alpha \left( \norm{\boldsymbol \mu - \boldsymbol \mu_\mathrm{pr}}_\mathrm{\Gamma_\mathrm{pr}^\mathrm{-1}}^\mathrm{2} +\mathrm{tr}\left( \Gamma_\mathrm{pr}^\mathrm{-1} \Gamma  \right)  \right) \Bigr).  \label{eq:minProblem}
\end{gather}
For $\alpha = 0.5$, $UB_\mathrm{\alpha}(q_\mathrm{pro,\phi} )$ \eqref{eq:minProblem} coincides with \eqref{eq:upperBoundFinal}. For $\alpha \neq 0.5$, the two quantities differ and with $UB_\mathrm{\alpha}(q_\mathrm{pro,\phi} )$ we lose a clear link with the lower bound $LB_\mathrm{\alpha}(q_\mathrm{pro,\phi} )$ \eqref{eq:lowerBound}. Nonetheless, the theorem of the next section, stated for $\mathcal F$ affine, explicits a relationship between $(\mathbf u_\mathrm{MAP},\Gamma_\mathrm{Lap})$ and the stationary points of $UB_\mathrm{\alpha}(q_\mathrm{pro,\phi} )$.

\emph{Remark.} While the minimum of the $UB_\mathrm{\alpha}(q_\mathrm{pro,\phi})$ \eqref{eq:minProblem} may not match the minimum of $LB_\mathrm{\alpha}(q_\mathrm{pro,\phi})$ \eqref{eq:lowerBound}, a relationship between the two minima will be discussed in the next section.

\emph{Remark.} In practice, to ensure that $\Gamma$ is SPD, we will use its lower triangular Cholesky factor $C \in \R^\mathrm{D \times D}$ with positive diagonal entries, such that $\Gamma = C C^\mathrm{T}$.

\subsection{Upper bound stationary points} \label{sec:anRes}
We now prove that the stationary points of \eqref{eq:minProblem} are related to $\mathbf u_\mathrm{MAP}$ and $\Gamma_\mathrm{Lap}$ when $\mathcal F$ is affine and provide an expression for $(\boldsymbol \mu_\mathrm{post},\Gamma_\mathrm{post})$. All the assumptions used to derive \eqref{eq:minProblem} are included in the hypotheses of the following theorem. Moreover, as discussed in \Cref{sec:buildingLossFunction}, we assume that all covariance matrices are SPD. We focus on the terms of \eqref{eq:minProblem} depending on $(\boldsymbol \mu,\Gamma)$, as the remaining terms do not influence the stationary points.
\begin{thm} \label{thm:convergence}
Let us assume that $\mathcal F(\mathbf u) = F \mathbf u + \mathbf f$, where $F \in \R^\mathrm{O\times D}, \mathbf f \in \R^\mathrm{O}$, and $\mathbf y \in \R^\mathrm{O}$. Assume Gaussian prior and noise models $\mathcal N(\boldsymbol \mu_\mathrm{pr}, \Gamma_\mathrm{pr})$ and $\mathcal N (\boldsymbol \mu_\mathrm{E}, \Gamma_\mathrm{E})$, respectively, and that the two random variables are independent. Then the posterior distribution $p_\mathrm{U|Y}(\mathbf u| \mathbf y) = \mathcal N (\mathbf u_\mathrm{MAP}, \Gamma_\mathrm{Lap})$, where:
\begin{align}
\mathbf u_\mathrm{MAP} &= \Gamma_\mathrm{Lap} \left( F^\mathrm{T} \Gamma_\mathrm{E}^\mathrm{-1} \left(  \mathbf y - \mathbf f - \boldsymbol \mu_\mathrm{E} \right) +\Gamma_\mathrm{pr}^\mathrm{-1} \boldsymbol \mu_\mathrm{pr}  \right) ,\label{eq:mapMuAffine}\\
\Gamma_\mathrm{Lap} &= \left( F^\mathrm{T} \Gamma_\mathrm{E}^\mathrm{-1}F + \Gamma_\mathrm{pr}^\mathrm{-1} \right)^\mathrm{-1}.\label{eq:mapGammaAffine}
\end{align}

Let $\alpha \in \left(0,1 \right)$ and suppose that the pdf $q_\mathrm{pro,\phi}(\mathbf u | \mathbf y)$ is $\mathcal N(\boldsymbol \mu, \Gamma)$, where $\Gamma = CC^\mathrm{T}$ and $C$ is a lower triangular matrix with positive diagonal entries. Let  $(\hat {\boldsymbol \mu}, \hat{C})$ (with $\hat{\Gamma} = \hat{C}\hat{C}^\mathrm{T}$) denote the stationary points of the loss function
\begin{align}
 L_\mathrm{\alpha,aff}(\boldsymbol \mu,\Gamma)  =&(1-\alpha) \left( \norm{\boldsymbol \mu - \boldsymbol \mu_\mathrm{pr}}_\mathrm{\Gamma^\mathrm{-1}}^\mathrm{2} + \mathrm{tr}\left(\Gamma^\mathrm{-1}\Gamma_\mathrm{pr}  \right) \right)\\
&+ \alpha \left(  \norm{\mathbf y - \boldsymbol \mu_\mathrm{E} - F \boldsymbol \mu -\mathbf f }_\mathrm{\Gamma_\mathrm{E}^\mathrm{-1}}^\mathrm{2} + \mathrm{tr} \left( \Gamma_\mathrm{E}^\mathrm{-1} F\Gamma F^\mathrm{T} \right) \right)  \\
 &+\alpha \left( \norm{\boldsymbol \mu - \boldsymbol \mu_\mathrm{pr}}_\mathrm{\Gamma_\mathrm{pr}^\mathrm{-1}}^\mathrm{2} +\mathrm{tr}\left( \Gamma_\mathrm{pr}^\mathrm{-1} \Gamma  \right)  \right). \label{eq:minProblemAffine2}
\end{align}
Define
\begin{align}
\boldsymbol \mu_\mathrm{post} &= \frac{1-\alpha}{\alpha} \Gamma_\mathrm{Lap} \hat{\Gamma}^\mathrm{-1} (\hat{\boldsymbol \mu}-\boldsymbol \mu_\mathrm{pr}) + \hat{\boldsymbol \mu}, \label{eq:muMap2}\\
\Gamma_\mathrm{post} &= \hat{\Gamma} A^\mathrm{-1} \hat{\Gamma}, \label{eq:gammaMap2}
\end{align}
where
\begin{align}
A =  \frac{1-\alpha}{\alpha}\left( (\hat{\boldsymbol \mu}- \boldsymbol \mu_\mathrm{pr})(\hat{\boldsymbol \mu}- \boldsymbol \mu_\mathrm{pr})^\mathrm{T} + \Gamma_\mathrm{pr} \right). \label{eq:Amatrix}
\end{align}
Then $(\boldsymbol \mu_\mathrm{post},\Gamma_\mathrm{post}) = (\mathbf u_\mathrm{MAP},\Gamma_\mathrm{Lap})$.\\
Moreover, $L_\mathrm{\alpha,aff}(\boldsymbol \mu,\Gamma)$ is separately convex in $\boldsymbol \mu$ and $\Gamma$.
\end{thm}

\begin{proof}
The posterior distribution is $\mathcal N (\mathbf u_\mathrm{MAP}, \Gamma_\mathrm{Lap})$ by Theorem 6.20 of \cite{stuart2010inverse}.

Deriving $L_\mathrm{\alpha,aff}$ with respect to $\boldsymbol \mu$ and leveraging the symmetry of the covariance matrices, we find:
\begin{align}
 \frac{\partial L_\mathrm{\alpha,aff}}{\partial \boldsymbol \mu} =& 2(1-\alpha)\Gamma^\mathrm{-1}(\boldsymbol \mu- \boldsymbol \mu_\mathrm{pr})\\
 &-2\alpha F^\mathrm{T} \Gamma_\mathrm{E}^\mathrm{-1}(\mathbf y - \boldsymbol \mu_\mathrm{E} - F \boldsymbol \mu -\mathbf f )\\
 & +2\alpha \Gamma_\mathrm{pr}^\mathrm{-1}(\boldsymbol \mu- \boldsymbol \mu_\mathrm{pr}).
\end{align}
Using  \eqref{eq:mapMuAffine} and \eqref{eq:mapGammaAffine}, we get:
\begin{align}
\frac{\partial L_\mathrm{\alpha,aff}}{\partial \boldsymbol \mu} = 2(1-\alpha)\Gamma^\mathrm{-1}(\boldsymbol \mu- \boldsymbol \mu_\mathrm{pr})+2\alpha \Gamma_\mathrm{Lap}^\mathrm{-1}(\boldsymbol \mu- \mathbf u_\mathrm{MAP}). \label{eq:lossMu}
\end{align}
Setting $\frac{\partial L_\mathrm{\alpha,aff}}{\partial \boldsymbol \mu} = \mathbf 0$ gives:
\begin{align}
\mathbf u_\mathrm{MAP} = \frac{1-\alpha}{\alpha} \Gamma_\mathrm{Lap}\hat{\Gamma}^\mathrm{-1}(\hat{\boldsymbol \mu}- \boldsymbol \mu_\mathrm{pr}) + \hat{\boldsymbol \mu} = \boldsymbol \mu_\mathrm{post}. \label{eq:stationaryMu}
\end{align}
Notice that we can divide by $\alpha$ since $\alpha > 0$. 

Similarly, deriving $L_\mathrm{\alpha,aff}$ with respect to $\Gamma$ and using the symmetry of the covariance matrices, we get:
\begin{align}
\frac{\partial L_\mathrm{\alpha,aff}}{\partial \Gamma} =& 2(1-\alpha)\left( - \Gamma^\mathrm{-1} (\boldsymbol \mu- \boldsymbol \mu_\mathrm{pr}) (\boldsymbol \mu- \boldsymbol \mu_\mathrm{pr})^\mathrm{T}\Gamma^\mathrm{-1} - \Gamma^\mathrm{-1} \Gamma_\mathrm{pr} \Gamma^\mathrm{-1}\right) \\
& + 2\alpha \left( F^\mathrm{T}\Gamma_\mathrm{E}^\mathrm{-1}F + \Gamma_\mathrm{pr}^\mathrm{-1}\right). \label{eq:lossGamma}
\end{align}
Setting $\frac{\partial L_\mathrm{\alpha,aff}}{\partial \Gamma} = 0$, we obtain:
\begin{align}
\Gamma_\mathrm{Lap} = \hat{\Gamma} A^\mathrm{-1} \hat{\Gamma} = \Gamma_\mathrm{post}.
\end{align}
Notice that $A^\mathrm{-1}$ is well-defined since, for $\alpha \in \left( 0,1 \right)$, $A$ is a positive multiple of the sum of a positive definite matrix ($\Gamma_\mathrm{pr}$) and a positive semidefinite matrix, therefore $A$ is a positive definite matrix. Therefore, also $ \hat{\Gamma} A^\mathrm{-1} \hat{\Gamma}$ is SPD as $\Gamma_\mathrm{Lap}$.

We now analyze the convexity of $L_\mathrm{\alpha,aff}(\boldsymbol \mu, \Gamma)$. Computing the Hessian of $L_\mathrm{\alpha,aff}$ with respect to $\boldsymbol \mu$, we obtain:
\begin{align}
 \frac{\partial^\mathrm{2} L_\mathrm{\alpha,aff}}{\partial \boldsymbol \mu^\mathrm{2}} = 2(1-\alpha)\Gamma^\mathrm{-1}+2\alpha F^\mathrm{T} \Gamma_\mathrm{E}^\mathrm{-1} F + 2 \alpha \Gamma_\mathrm{pr}^\mathrm{-1}.
\end{align}
The Hessian is positive definite, being the sum of two positive definite matrices and a positive semidefinite matrix. Therefore, $L_\mathrm{\alpha,aff}$ is convex in $\boldsymbol \mu$.

We now analyze the convexity in $\Gamma$. Defining the SPD matrix $B = \Gamma_\mathrm{pr}+(\boldsymbol \mu - \boldsymbol \mu_\mathrm{pr})(\boldsymbol \mu - \boldsymbol \mu_\mathrm{pr})^\mathrm{T}$, we have:
\begin{align}
\norm{\boldsymbol \mu - \boldsymbol \mu_\mathrm{pr}}_\mathrm{\Gamma^\mathrm{-1}}^\mathrm{2} + \mathrm{tr}\left( \Gamma^\mathrm{-1}\Gamma_\mathrm{pr} \right) = \mathrm{tr}\left( \Gamma^\mathrm{-1}B\right).
\end{align}
For $\lambda \in [0,1]$ and SPD matrices $\Gamma_\mathrm{1}$ and $\Gamma_\mathrm{2}$, it holds that
\begin{align}
\lambda \Gamma_\mathrm{1}^\mathrm{-1}+(1-\lambda) \Gamma_\mathrm{2}^\mathrm{-1} - 
(\lambda \Gamma_\mathrm{1} + (1-\lambda) \Gamma_\mathrm{2})^\mathrm{-1}
\end{align} 
is positive semi-definite. Consequently,
\begin{align}
\mathrm{tr}\left((\lambda \Gamma_\mathrm{1}^\mathrm{-1}+(1-\lambda) \Gamma_\mathrm{2}^\mathrm{-1} - 
(\lambda \Gamma_\mathrm{1} + (1-\lambda) \Gamma_\mathrm{2})^\mathrm{-1}) B\right)=\\
\mathrm{tr}\left(B^\mathrm{\frac{1}{2}}(\lambda \Gamma_\mathrm{1}^\mathrm{-1}+(1-\lambda) \Gamma_\mathrm{2}^\mathrm{-1} - 
(\lambda \Gamma_\mathrm{1} + (1-\lambda) \Gamma_\mathrm{2})^\mathrm{-1}) B^\mathrm{\frac{1}{2}}\right)\ge 0,
\end{align}
because $B^\mathrm{\frac{1}{2}}(\lambda \Gamma_\mathrm{1}^\mathrm{-1}+(1-\lambda) \Gamma_\mathrm{2}^\mathrm{-1} - 
(\lambda \Gamma_\mathrm{1} + (1-\lambda) \Gamma_\mathrm{2})^\mathrm{-1}) B^\mathrm{\frac{1}{2}}$ is symmetric positive semidefinite. As a result,
\begin{align}
\mathrm{tr}\left((\lambda \Gamma_\mathrm{1} + (1-\lambda) \Gamma_\mathrm{2})^\mathrm{-1} B\right) \le \mathrm{tr}\left((\lambda \Gamma_\mathrm{1}^\mathrm{-1}+(1-\lambda) \Gamma_\mathrm{2}^\mathrm{-1}) B\right) = \lambda \mathrm{tr}\left( \Gamma_\mathrm{1}^\mathrm{-1}B\right)+(1-\lambda) \mathrm{tr}\left( \Gamma_\mathrm{2}^\mathrm{-1}B\right),
\end{align}
showing that $\mathrm{tr}\left( \Gamma^\mathrm{-1}B\right)$ is convex.\\
The remaining terms of $L_\mathrm{\alpha,aff}$ are affine in $\Gamma$ and hence convex. As the sum of convex functions is convex, it follows that $L_\mathrm{\alpha,aff}$ is convex in $\Gamma$.
\end{proof}
 
 \emph{Remark.} Expressions \eqref{eq:mapMuAffine}, \eqref{eq:mapGammaAffine} and \eqref{eq:minProblemAffine2} derive from \eqref{eq:mapMu}, \eqref{eq:mapGamma} and \eqref{eq:minProblem}, respectively, when $\mathcal F$ is affine.

In the general case when $\mathcal F$ is not affine, the expected value of \eqref{eq:minProblem} cannot be computed explicitly. Instead, we estimate it using the sample mean:
\begin{align}
& \E_\mathrm{q_\mathrm{pro,\phi}(\mathbf u | \mathbf y)}\left[ \norm{\mathbf y - \boldsymbol \mu_\mathrm{E} - \mathcal F(\mathbf u)}_\mathrm{\Gamma_\mathrm{E}^\mathrm{-1}}^\mathrm{2} \right] \sim \frac{1}{K} \sum_\mathrm{k = 1}^\mathrm{K} \norm{\mathbf y - \boldsymbol \mu_\mathrm{E} - \mathcal F(\mathbf u^\mathrm{k})}_\mathrm{\Gamma_\mathrm{E}^\mathrm{-1}}^\mathrm{2}, \label{eq:estMean}
\end{align}
where $K \in \N $ is the number of samples and, for $k=1,\dots,K$,  $\mathbf u^\mathrm{k} = \boldsymbol \mu + C \boldsymbol \epsilon^\mathrm{k}$ with $\boldsymbol  \epsilon^\mathrm{k} \in \R^\mathrm{D}$ and $\boldsymbol \epsilon^\mathrm{k} \sim \mathcal N(\mathbf 0, I)$. Here, each $\mathbf u^\mathrm{k}$ represents a sample drawn from $\mathcal N(\boldsymbol \mu, \Gamma)$. The sampling process results in slow convergence to the actual mean, making computationally expensive retrieving the result of \Cref{thm:convergence}. The corresponding loss function, used when the map $\mathcal F$ is known and not expensive to compute, becomes:
\begin{gather}
L_\mathrm{\alpha}(\boldsymbol \mu,\Gamma;\mathcal F) = (1-\alpha) \left( \norm{\boldsymbol \mu - \boldsymbol \mu_\mathrm{pr}}_\mathrm{\Gamma^\mathrm{-1}}^\mathrm{2} + \mathrm{tr}\left(\Gamma^\mathrm{-1}\Gamma_\mathrm{pr}  \right) \right) \\
 +\alpha \frac{1}{K} \sum_\mathrm{k = 1}^\mathrm{K} \norm{\mathbf y - \boldsymbol \mu_\mathrm{E} - \mathcal F(\mathbf u^\mathrm{k})}_\mathrm{\Gamma_\mathrm{E}^\mathrm{-1}}^\mathrm{2} \\
 +\alpha \left( \norm{\boldsymbol \mu - \boldsymbol \mu_\mathrm{pr}}_\mathrm{\Gamma_\mathrm{pr}^\mathrm{-1}}^\mathrm{2} +\mathrm{tr}\left( \Gamma_\mathrm{pr}^\mathrm{-1} \Gamma  \right)  \right). \label{eq:minProblem3}
\end{gather}

\subsection{Enhanced uncertainty quantification variational autoencoders} \label{sec:UQVAE}
\begin{figure}[t!]
	\centering
 	\includegraphics[width=\linewidth, height=6cm,keepaspectratio]{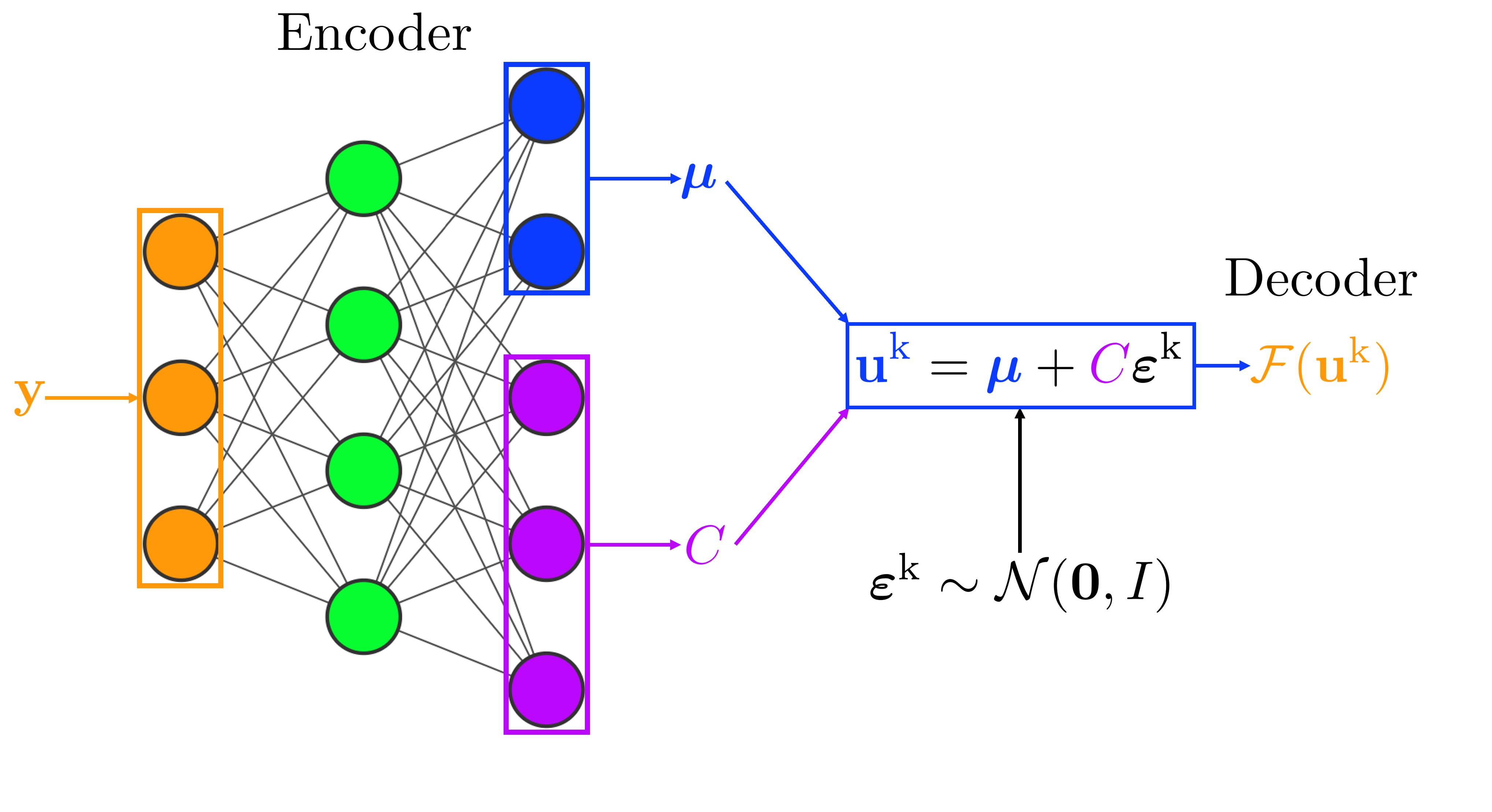}
	\caption{Scheme of the eUQ-VAE with the evaluation of $\mathcal F$. The observational data $\mathbf y \in \R^\mathrm{O}$ is passed to the encoder that computes the proxy distribution mean $\boldsymbol \mu \in \R^\mathrm{D}$ and Cholesky factor of the covariance matrix $C\in \R^\mathrm{D \times D}$. $K$ samples $\{\mathbf u^\mathrm{k}\}_\mathrm{k=1}^\mathrm{K}$ are drawn from the proxy distribution $q_\mathrm{pro,\phi}(\mathbf u| \mathbf y) =  \mathcal N(\boldsymbol \mu, CC^\mathrm{T})$ and their image through the decoder $\mathcal F$ is computed. $\{ {\boldsymbol \epsilon}^\mathrm{k} \}_\mathrm{k=1}^\mathrm{K}$ are drawn from a multivariate standard Gaussian distribution and are used to compute $\{\mathbf u^\mathrm{k}\}_\mathrm{k=1}^\mathrm{K}$ by means of $\boldsymbol \mu$ and $C$.}
	\label{fig:linearVAE}
\end{figure}

In this section, we introduce the eUQ-VAE framework, a semi-supervised model constrained learning approach. Let $\phi$ denote the weights and the biases of a feed forward neural network. Our aim is to investigate whether \Cref{thm:convergence} still holds when $(\boldsymbol \mu, \Gamma)$ are computed using a neural network. Specifically, we employ a fully connected neural network. As illustrated in \Cref{fig:linearVAE}, the architecture of the eUQ-VAE consists of a neural network serving as the encoder and of $\mathcal F$ acting as the decoder. The eUQ-VAE architecture coincides with that of the UQ-VAE \cite{goh2021solving}. The difference in the two approaches lies in the choice of the loss function, which alone enhances the approximation of the posterior mean and covariance. 

\emph{Remark.} Unlike approaches for deterministic inverse problems (e.g., TAEN \cite{nguyen5081218taen}), the latent space of the autoencoder is composed of the proxy mean and the Choleksy factor of the covariance matrix, rather than estimates of the actual model parameters.

\begin{cor} \label{cor:convergenceNN}
Let $L_\mathrm{\alpha,aff}(\phi) = L_\mathrm{\alpha,aff}(\boldsymbol \mu(\phi), \Gamma(\phi))$ and suppose the assumptions of  \Cref{thm:convergence} hold. Consider a single-layer linear neural network such that:
\begin{align}
\boldsymbol \mu &= W_\mathrm{\boldsymbol \mu} \mathbf y + \mathbf b_\mathrm{\boldsymbol \mu}, \label{eq:muPost}\\
C &= \mathrm{vec}_\mathrm{L}^\mathrm{-1}(\mathbf l) + \mathrm{diag}(\boldsymbol \sigma), \label{eq:cPost}\\
\boldsymbol \sigma &= \mathrm{exp} \left(  W_\mathrm{\boldsymbol \sigma} \mathbf y + \mathbf b_\mathrm{\boldsymbol \sigma} \right), \label{eq:sigma}\\
\mathbf l &= W_\mathrm{\mathbf l} \mathbf y + \mathbf b_\mathrm{\mathbf l}, \label{eq:l}
\end{align}
where $W_\mathrm{\boldsymbol \mu} \in \R^\mathrm{D \times O}$ and $W_\mathrm{\boldsymbol \sigma} \in \R^\mathrm{D \times O}, W_\mathrm{\mathbf l} \in \R^\mathrm{\frac{D(D-1)}{2} \times O}$ are matrices representing the weights of the neural network, $\mathbf b_\mathrm{\boldsymbol \mu} \in \R^\mathrm{D}, \mathbf b_\mathrm{\boldsymbol \sigma} \in \R^\mathrm{D}$ and $\mathbf b_\mathrm{\mathbf l} \in \R^\mathrm{\frac{D(D-1)}{2}}$ are the biases of the output layer. The functions $\mathrm{exp}$, $\mathrm{vec}_\mathrm{L}$ and $\mathrm{diag}$ are the elementwise exponential, the vectorization of a strictly lower triangular matrix by rows and the construction of a diagonal matrix from its vector diagonal, respectively.\\
The stationary points $\hat \phi$ of $L_\mathrm{\alpha,aff}(\phi)$ satisfy $(\boldsymbol \mu_\mathrm{post}(\hat \phi),\Gamma_\mathrm{post}(\hat \phi)) = (\mathbf u_\mathrm{MAP},\Gamma_\mathrm{Lap})$.
\end{cor}
\begin{proof}
The proof follows from the chain rule. Recall that for $i,j = 1,\dots D$:
\begin{align}
\mathrm{vec}_\mathrm{L}^\mathrm{-1}(\mathbf l)_\mathrm{i,j} =
\begin{cases}
0 \quad &\text{if } j \ge i,\\
l_\mathrm{i-1+j-1+\sum_\mathrm{k=1}^\mathrm{i-3}k} = l_\mathrm{i+j-2+\frac{(i-3)(i-2)}{2}}&\text{if } j < i. \label{eq:bijection}
\end{cases}
\end{align}
We explicitly write the components of $\boldsymbol \mu$ and $C$ element-wise:
\begin{gather}
 \mu_\mathrm{i} = \sum_\mathrm{k=1}^\mathrm{O} (W_\mathrm{\boldsymbol \mu})_\mathrm{i,k}y_\mathrm{k}+(b_\mathrm{\boldsymbol \mu})_\mathrm{i},\\
C_\mathrm{i,j} = 
\begin{cases}
0 \quad &\text{if } j>i,\\
\text{exp}\left(\sum_\mathrm{k=1}^\mathrm{O} (W_\mathrm{\boldsymbol \sigma})_\mathrm{i,k}y_\mathrm{k}+(b_\mathrm{\boldsymbol \sigma})_\mathrm{i} \right) & \text{if } j=i,\\
\sum_\mathrm{k=1}^\mathrm{O} (W_\mathrm{\mathbf l})_\mathrm{i+j-2+\frac{(i-3)(i-2)}{2},k}y_\mathrm{k}+(b_\mathrm{\mathbf l})_\mathrm{i+j-2+\frac{(i-3)(i-2)}{2}} & \text{if } j<i.
\end{cases}
\end{gather}
We recall that:
\begin{align} 
\frac{\partial \Gamma_\mathrm{i,j}}{\partial C_\mathrm{k,l}} = \delta_\mathrm{k,i}C_\mathrm{j,l}+\delta_\mathrm{k,j}C_\mathrm{i,l},
\end{align}
where $\delta_\mathrm{\cdot,\cdot}$ denotes the Kronecker delta. The derivatives of $L_\mathrm{\alpha,aff}(\phi)$ with respect to the encoder biases are computed as follows:
\begin{align}
\frac{\partial L_\mathrm{\alpha,aff}}{\partial (b_\mathrm{\boldsymbol \mu})_\mathrm{d}} = &
\sum_\mathrm{i=1}^\mathrm{D} \frac{\partial L_\mathrm{\alpha,aff}}{\partial \mu_\mathrm{i}}\frac{\partial  \mu_\mathrm{i}}{\partial (b_\mathrm{\boldsymbol \mu})_\mathrm{d}} 
=\sum_\mathrm{i=1}^\mathrm{D} \frac{\partial L_\mathrm{\alpha,aff}}{\partial \mu_\mathrm{i}} \delta_\mathrm{i,d} =
 \frac{\partial L_\mathrm{\alpha,aff}}{\partial \mu_\mathrm{d}},\\
\frac{\partial L_\mathrm{\alpha,aff}}{\partial (b_\mathrm{\boldsymbol \sigma})_\mathrm{d}} 
=&\sum_\mathrm{i,j=1}^\mathrm{D} \frac{\partial L_\mathrm{\alpha,aff}}{\partial \Gamma_\mathrm{i,j}}\frac{\partial  \Gamma_\mathrm{i,j}}{\partial (b_\mathrm{\boldsymbol \sigma})_\mathrm{d}}
= \sum_\mathrm{i,j=1}^\mathrm{D} \frac{\partial L_\mathrm{\alpha,aff}}{\partial \Gamma_\mathrm{i,j}} \sum_\mathrm{k,l=1}^\mathrm{D}\frac{\partial  \Gamma_\mathrm{i,j}}{\partial C_\mathrm{k,l}} \frac{\partial C_\mathrm{k,l}}{\partial (b_\mathrm{\boldsymbol \sigma})_\mathrm{d}} =\\
&\sum_\mathrm{i,j=1}^\mathrm{D} \frac{\partial L_\mathrm{\alpha,aff}}{\partial \Gamma_\mathrm{i,j}} \sum_\mathrm{k,l=1}^\mathrm{D}\frac{\partial  \Gamma_\mathrm{i,j}}{\partial C_\mathrm{k,l}} C_\mathrm{k,k}\delta_\mathrm{k,l}\delta_\mathrm{k,d}=
\sum_\mathrm{i,j=1}^\mathrm{D} \frac{\partial L_\mathrm{\alpha,aff}}{\partial \Gamma_\mathrm{i,j}} \frac{\partial  \Gamma_\mathrm{i,j}}{\partial C_\mathrm{d,d}} C_\mathrm{d,d} =\\
& C_\mathrm{d,d} \sum_\mathrm{i,j=1}^\mathrm{D} \frac{\partial L_\mathrm{\alpha,aff}}{\partial \Gamma_\mathrm{i,j}} (\delta_\mathrm{d,i}C_\mathrm{j,d}+\delta_\mathrm{d,j}C_\mathrm{i,d}) = 2C_\mathrm{d,d} \sum_\mathrm{i=1}^\mathrm{D}  \frac{\partial L_\mathrm{\alpha,aff}}{\partial \Gamma_\mathrm{d,i}}C_\mathrm{i,d} = 2C_\mathrm{d,d}\left(\frac{\partial L_\mathrm{\alpha,aff}}{\partial \Gamma} C \right)_\mathrm{d,d},
\end{align}
for $d=1,\dots,D$. To compute the derivatives with respect to $\mathbf b_\mathrm{\mathbf l}$, we exploit  the bijection induced by the $\mathrm{vec}_\mathrm{L}$ function between the sets $\{(2,1),(3,1),(3,2),\dots,(D,D-1)\}$ and $\{1,2,\dots,D(D-1)/2\}$, associating the strictly lower triangular matrix indices of the nonzero coefficients with the indices of a $D(D-1)/2$ dimensional vector \eqref{eq:bijection}. By a computation analogous to that for $\partial L_\mathrm{\alpha,aff}/\partial (b_\mathrm{\boldsymbol \sigma})_\mathrm{d}$, we obtain:
\begin{align}
\frac{\partial L_\mathrm{\alpha,aff}}{\partial (b_\mathrm{\mathbf l})_\mathrm{m+n-2+\frac{(m-3)(m-2)}{2}}}=&
\sum_\mathrm{i,j=1}^\mathrm{D} \frac{\partial L_\mathrm{\alpha,aff}}{\partial \Gamma_\mathrm{i,j}}\sum_\mathrm{k,l=1}^\mathrm{D}\frac{\partial  \Gamma_\mathrm{i,j}}{\partial C_\mathrm{k,l}} \frac{\partial C_\mathrm{k,l}}{\partial (b_\mathrm{\mathbf l})_\mathrm{m+n-2+\frac{(m-3)(m-2)}{2}}} = \\
& \sum_\mathrm{i,j=1}^\mathrm{D} \frac{\partial L_\mathrm{\alpha,aff}}{\partial \Gamma_\mathrm{i,j}}\sum_\mathrm{k,l=1}^\mathrm{D}\frac{\partial  \Gamma_\mathrm{i,j}}{\partial C_\mathrm{k,l}} \delta_\mathrm{k,m} \delta_\mathrm{l,n} = 2\left(\frac{\partial L_\mathrm{\alpha,aff}}{\partial \Gamma} C \right)_\mathrm{m,n} \; \text{for } 1\le n <m \le D.
\end{align}
Setting these derivatives to zero yields the stationary conditions:
\begin{align}
&\frac{\partial{L_\mathrm{\alpha,aff}}}{\partial \boldsymbol \mu} = \mathbf 0,\\
&\left(\frac{\partial L_\mathrm{\alpha,aff}}{\partial \Gamma} C \right)_\mathrm{m,n}  = 0\quad \text{for } 1 \le n \le m \le D. \label{eq:condStat}
\end{align}
From \eqref{eq:condStat}, we can identify the following conditions:
\begin{align}
\frac{\partial L_\mathrm{\alpha,aff}}{\partial \Gamma_\mathrm{i,1:i}}^\mathrm{T} C_\mathrm{1:i,1:i} = \mathbf 0 \quad \text{for } i = 1,\dots,D,
\end{align}
where $1:i$ denotes rows or columns from the first to the $i$-th. Since $C$ is lower triangular with positive diagonal entries, each $C_\mathrm{1:i,1:i}$ is invertible, leading to:
\begin{align}
\frac{\partial L_\mathrm{\alpha,aff}}{\partial \Gamma_\mathrm{i,1:i}} = \mathbf 0 \quad \text{for } i = 1,\dots,D.
\end{align}
Hence, the lower triangular part of $\partial L_\mathrm{\alpha,aff}/\partial \Gamma$ is zero. Because $\Gamma$ is symmetric, the Jacobian is also symmetric, giving
\begin{align}
\frac{\partial{L_\mathrm{\alpha,aff}}}{\partial \Gamma} = 0,
\end{align}
which satisfies the stationary condition \eqref{eq:condStat}.\\
Finally, the derivatives of $L_\mathrm{\alpha,aff}(\phi)$ with respect to the encoder weights are:
\begin{align}
\frac{\partial L_\mathrm{\alpha,aff}}{\partial  (W_\mathrm{\boldsymbol \mu})_\mathrm{d,o}}  
=\sum_\mathrm{i=1}^\mathrm{D} \frac{\partial L_\mathrm{\alpha,aff}}{\partial \mu_\mathrm{i}}\frac{\partial  \mu_\mathrm{i}}{\partial (W_\mathrm{\boldsymbol \mu})_\mathrm{d,o}} = 0,\\
 \frac{\partial L_\mathrm{\alpha,aff}}{\partial (W_\mathrm{\boldsymbol \sigma})_\mathrm{d,o}} 
=\sum_\mathrm{i,j=1}^\mathrm{D} \frac{\partial L_\mathrm{\alpha,aff}}{\partial \Gamma_\mathrm{i,j}}\frac{\partial  \Gamma_\mathrm{i,j}}{\partial (W_\mathrm{\boldsymbol \sigma})_\mathrm{d,o}}  = 0,\\
 \frac{\partial L_\mathrm{\alpha,aff}}{\partial (W_\mathrm{\mathbf l})_\mathrm{t,o}} 
=\sum_\mathrm{i,j=1}^\mathrm{D} \frac{\partial L_\mathrm{\alpha,aff}}{\partial \Gamma_\mathrm{i,j}}\frac{\partial  \Gamma_\mathrm{i,j}}{\partial (W_\mathrm{\mathbf l})_\mathrm{t,o}}= 0,
\end{align}
for $d=1,\dots,D$, $o = 1,\dots,O$ and $t=1,\dots,D(D-1)/2$.\\
From \Cref{thm:convergence}, it follows that $(\boldsymbol \mu_\mathrm{post}(\hat \phi),\Gamma_\mathrm{post}(\hat \phi)) = (\mathbf u_\mathrm{MAP},\Gamma_\mathrm{Lap})$.
\end{proof}
 
This corollary guarantees that, if a minimum $(\hat {\boldsymbol \mu}, \hat{\Gamma})$ of the loss function \eqref{eq:minProblemAffine2} exists, the training process of the neural network attempts to enforce the relationships \eqref{eq:muMap2} and \eqref{eq:gammaMap2}.

If the neural network has nonlinear activation functions and contains more than one layer, $(\boldsymbol \mu,C)$ are computed as in \eqref{eq:muPost}, \eqref{eq:cPost}, \eqref{eq:sigma} and \eqref{eq:l}, using the outputs of the network's last layer in place of $W_\mathrm{\boldsymbol \mu}\mathbf y + \mathbf b_\mathrm{\boldsymbol \mu}$, $W_\mathrm{\boldsymbol \sigma}\mathbf y + \mathbf b_\mathrm{\boldsymbol \sigma}$ and $W_\mathrm{\mathbf l}\mathbf y + \mathbf b_\mathrm{\mathbf l}$.

\begin{figure}[t!]
	\centering
 	\includegraphics[width=\linewidth, height=6cm,keepaspectratio]{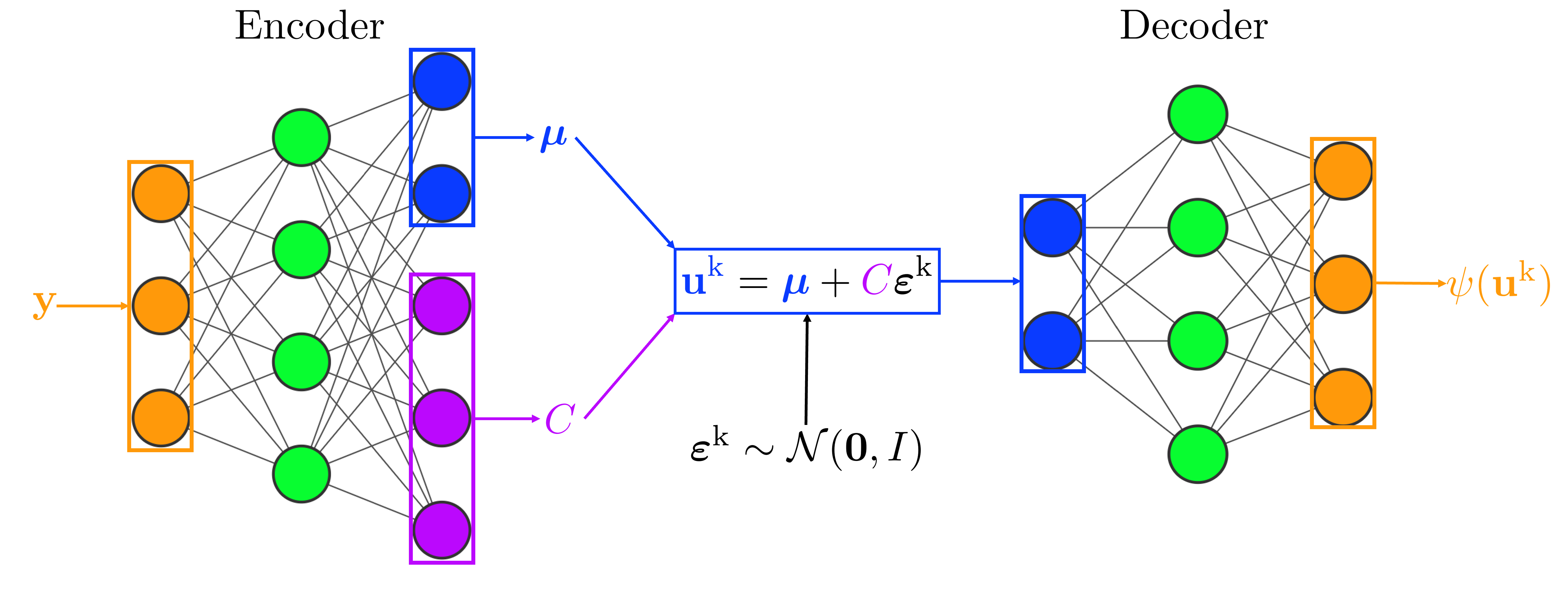}
	\caption{Scheme of the eUQ-VAE with a neural network $\psi$ in place of the evaluation of $\mathcal F$. The observational data $\mathbf y \in \R^\mathrm{O}$ is passed to the encoder that computes the proxy distribution mean $\boldsymbol \mu \in \R^\mathrm{D}$ and Cholesky factor of the covariance matrix $C\in \R^\mathrm{D \times D}$. $K$ samples $\{\mathbf u^\mathrm{k}\}_\mathrm{k=1}^\mathrm{K}$ are drawn from the proxy distribution $q_\mathrm{pro,\phi}(\mathbf u| \mathbf y) =  \mathcal N(\boldsymbol \mu, CC^\mathrm{T})$ and their image through the decoder $\psi$ is computed. $\{ {\boldsymbol \epsilon}^\mathrm{k} \}_\mathrm{k=1}^\mathrm{K}$ are drawn from a multivariate standard Gaussian distribution and are used to compute $\{\mathbf u^\mathrm{k}\}_\mathrm{k=1}^\mathrm{K}$ by means of $\boldsymbol \mu$ and $C$.}
	\label{fig:VAE}
\end{figure}

If $\mathcal F$ is unknown or computationally expensive to evaluate, it can be approximated using a second neural network $\psi$ acting as the decoder (\Cref{fig:VAE}) \cite{goh2021solving}. By substituting this approximation in \eqref{eq:minProblem3}, the resulting loss function becomes:
\begin{gather}
L_\mathrm{\alpha}(\boldsymbol \mu,\Gamma;\psi) =  (1-\alpha) \left(\norm{\boldsymbol \mu - \boldsymbol \mu_\mathrm{pr}}_\mathrm{\Gamma^\mathrm{-1}}^\mathrm{2} + \mathrm{tr}\left(\Gamma^\mathrm{-1}\Gamma_\mathrm{pr}  \right) \right) \\
+\alpha \frac{1}{K} \sum_\mathrm{k = 1}^\mathrm{K} \norm{\mathbf y - \boldsymbol \mu_\mathrm{E} - \psi(\mathbf u^\mathrm{k})}_\mathrm{\Gamma_\mathrm{E}^\mathrm{-1}}^\mathrm{2}\\
 +\alpha \left( \norm{\boldsymbol \mu - \boldsymbol \mu_\mathrm{pr}}_\mathrm{\Gamma_\mathrm{pr}^\mathrm{-1}}^\mathrm{2} +\mathrm{tr}\left( \Gamma_\mathrm{pr}^\mathrm{-1} \Gamma  \right)  \right).\label{eq:minProblem4}
\end{gather}
The simultaneous training of the encoder and the decoder could lead to unexpected results as both networks compete to minimize the loss function. Therefore, the decoder is trained before to the encoder. This also keeps a parallelism between the loss function \eqref{eq:minProblem4} and \eqref{eq:minProblem3}, where the map $\mathcal F$ from parameters to observational data is fixed. More theoretical insights can be analyzed by minimizing the loss function with respect to both the encoder and decoder weights and biases \cite{nguyen5081218taen}.

After the training of the eUQ-VAE, we retain only the encoder component which is used to compute $(\boldsymbol \mu_\mathrm{post},C_\mathrm{post})$.

In the previous theoretical analysis, we considered the case of a single observational data $\mathbf y$. When extending to $M\in \N$ samples, the loss functions \eqref{eq:minProblem3} and \eqref{eq:minProblem4} are replaced by the mean of the individual losses across all the samples.

\subsubsection{Data normalization and weights and biases initialization}\label{sec:DN_WI}
Before the training, it is a good practice to normalize the data to improve the generalization capabilities of the neural network. In our case, normalization affects also the probability distributions. Consider a dataset $\{(\mathbf u^\mathrm{(m)}, \mathbf y^\mathrm{(m)})\}_\mathrm{m=1}^\mathrm{M}$ with $M \in \N$, where $\mathbf y^\mathrm{(m)} = \mathcal F(\mathbf u^\mathrm{(m)}) + \boldsymbol \epsilon^\mathrm{(m)}$ and $\mathbf u^\mathrm{(m)}$ and $\boldsymbol \epsilon^\mathrm{(m)}$ are sampled from $\mathcal N(\boldsymbol \mu_\mathrm{pr},\Gamma_\mathrm{pr})$ and $\mathcal N(\boldsymbol \mu_\mathrm{E},\Gamma_\mathrm{E})$, respectively.

Normalizing the data, for example to the range $[0,1]$, means that there exist vectors $\mathbf a, \mathbf b \in \R^\mathrm{D}$ and $\mathbf c, \mathbf d \in \R^\mathrm{O}$ such that $\forall \, m = 1, \dots ,M$:
\begin{align}
\bar{\mathbf u}^\mathrm{(m)} &= \mathbf u^\mathrm{(m)} \odot \mathbf a + \mathbf b \in [0,1]^\mathrm{D},\\
\bar{\mathbf y}^\mathrm{(m)} &= \mathbf y^\mathrm{(m)} \odot \mathbf c + \mathbf d \in [0,1]^\mathrm{O}, \label{eq:yNorm}
\end{align}
where $\odot$ denotes the Hadamard product. By the variables transformation and the model \eqref{eq:model}, we get:
\begin{align*}
\bar{\mathbf y}^\mathrm{(m)}= \mathbf y^\mathrm{(m)} \odot \mathbf c + \mathbf d = \mathcal F((\bar{\mathbf u}^\mathrm{(m)}-\mathbf b)\oslash \mathbf a)\odot \mathbf c +\mathbf d+\boldsymbol \epsilon^\mathrm{(m)} \odot \mathbf c = \bar{\mathcal F}(\bar{\mathbf u}^\mathrm{(m)})+\bar{\boldsymbol \epsilon }^\mathrm{(m)},
\end{align*}
where $\bar{\mathcal F} (\bar {\mathbf u}^\mathrm{(m)}) \coloneqq \mathcal F((\bar{\mathbf u}^\mathrm{(m)}-\mathbf b)\oslash \mathbf a) \odot \mathbf c + \mathbf d $ and $\bar {\boldsymbol \epsilon}^\mathrm{(m)} = \boldsymbol \epsilon^\mathrm{(m)} \odot \mathbf c$. As a result, an observational model analogous to \eqref{eq:model} holds for the normalized scales:
\begin{align}
\bar{Y} = \bar{\mathcal F}(\bar{U})+\bar{E},
\end{align}
where $\bar{Y}, \bar{U}$ and $\bar{E}$ are the observable data, the parameters and the noise random variables of the normalized mathematical model. Since $U \sim \mathcal N(\boldsymbol \mu_\mathrm{pr}, \Gamma_\mathrm{pr}), E \sim \mathcal N(\boldsymbol \mu_\mathrm{E}, \Gamma_\mathrm{E})$ and the normalization is an affine function, the normalized random variables $\bar U$ and $\bar E$ are also Gaussian random variables, with means and covariances:
\begin{align}
 \bar{\boldsymbol \mu}_\mathrm{pr} &= \boldsymbol \mu_\mathrm{pr} \odot \mathbf a + \mathbf b,\\
 \bar{\Gamma}_\mathrm{pr} &= \Gamma_\mathrm{pr} \odot (\mathbf a \mathbf a^\mathrm{T}),\\
 \bar{\boldsymbol \mu}_\mathrm{E} &= \boldsymbol \mu_\mathrm{E} \odot \mathbf c,\\
 \bar{\Gamma}_\mathrm{E} &= \Gamma_\mathrm{E} \odot (\mathbf c \mathbf c^\mathrm{T}). 
\end{align}

Moreover, $\mathcal F$ is affine if and only if $\bar{\mathcal F}$ is affine, then \Cref{thm:convergence} holds also for the normalized problem. Once we minimize \eqref{eq:minProblem3} or \eqref{eq:minProblem4} with respect to the normalized variables and obtain $(\bar{\boldsymbol \mu}_\mathrm{post},\bar{\Gamma}_\mathrm{post})$, we transform back to the original scale by:
\begin{align}
\boldsymbol \mu_\mathrm{post} &= (\bar{\boldsymbol \mu}_\mathrm{post}-\mathbf b) \oslash \mathbf a,\\
\boldsymbol \Gamma_\mathrm{post}& = \bar{\Gamma}_\mathrm{post} \odot ((\mathbf 1\oslash \mathbf a)(\mathbf 1\oslash \mathbf a)^\mathrm{T}),
\end{align}
where $\mathbf 1 \in \R^\mathrm{D}$ is a vector of ones.

The initialization of the weights and biases of the encoder plays a crucial role in its training. As a stationary point of the first and third rows of the loss function \eqref{eq:minProblem4} is achieved for $\boldsymbol \mu=\boldsymbol \mu_\mathrm{pr}$ and $C=\sqrt[4]{\frac{1-\alpha}{\alpha}}C_\mathrm{pr}$ (Cholesky factor of $\Gamma_\mathrm{pr}$), we initialize the network hyperparameters in such a way to obtain outputs approximating these values. The encoder weights are initialized according to the Xavier uniform initializer \cite{glorot2010understanding}. Then, the weights of the last layer are scaled by $10^\mathrm{-4}$. This ensures that, up to a small discrepancy, the output of the encoder is given by the biases of the last layer. All the encoder biases are set to $0$, except for those of the last layer, which are fixed to $\boldsymbol \mu_\mathrm{pr}$ and $\mathbf c_\mathrm{pr}$, where $\mathbf c_\mathrm{pr}$ is given by applying the inverse of the right hand side transformation of \eqref{eq:cPost} to $\sqrt[4]{\frac{1-\alpha}{\alpha}}C_\mathrm{pr}$.

\section{Numerical results}\label{sec:results}
In this section, we evaluate the performance of the proposed eUQ-VAE method: in \Cref{sec:test1}, we verify \Cref{thm:convergence} and compare the results of the eUQ-VAE with those of UQ-VAE \cite{goh2021solving}; in \Cref{sec:test2}, we analyze the generalization properties of eUQ-VAE relative to UQ-VAE and examine the behavior of $L_\mathrm{\alpha}(\boldsymbol \mu, \Gamma;\mathcal F)$ \eqref{eq:minProblem3}, the lower bounds $LB_\mathrm{\alpha}(q_\mathrm{pro,\phi})$, $LB_\mathrm{\alpha}(q_\mathrm{\phi})$ \eqref{eq:lowerBound} and the upper bound $UB_\mathrm{\alpha}(q_\mathrm{pro,\phi})$ \eqref{eq:minProblem} during eUQ-VAE training for $\alpha = 0.5$; in \Cref{sec:test3}, we apply eUQ-VAE, UQ-VAE and Markov Chains Monte Carlo (MCMC) to a Laplace equation.

For a fair comparison between the two NN approaches, we use identical architectures. In all experiments, we employ the ReLU activation function, RMSProp \cite{tieleman2012lecture} as optimizer, set $K=2^\mathrm{12}$ and normalize the data to the interval $[0,1]$. For brevity, this normalization will not be explicitly mentioned in the following sections.

To keep the computational cost of estimating the expectation of \eqref{eq:minProblem3} or \eqref{eq:minProblem4} manageable, we approximate it using Sobol' sequences \cite{sobol1967distribution}, a Quasi-Monte Carlo method that boosts the convergence of the sample mean to the expected value.

\subsection{Validation of the theoretical result} \label{sec:test1}
We assume that the parameters $U$ are distributed as a Gaussian random field over the interval $[-2,2]$. We evaluate the parameters at $D = 20$ equally spaced points $x_\mathrm{j}$. We set $\boldsymbol \mu_\mathrm{pr} = \mathbf 1$ and define the entries of $\Gamma_\mathrm{pr}$ as:
\begin{align}
\left( \Gamma_\mathrm{pr} \right)_\mathrm{i,j} = \frac{\mathrm{e}^\mathrm{-\beta |x_\mathrm{j}-x_\mathrm{k} |}}{2\beta \gamma},
\end{align}
where $\beta = 1/5$ and $\gamma = 1/8$. This choice of $\Gamma_\mathrm{pr}$, together with regularity conditions on $\mathcal F$, ensures the well-posedness of general $1$-D Bayesian inverse problems \cite{stuart2010inverse,daon2018mitigating}. Here, the dimension refers to the domain where the parameter random variable is defined.\\
We consider a linear function represented by a matrix $F \in \R^\mathrm{O \times D}$ with $O = 15$. The entries of the matrix are sampled form a uniform distribution over the interval $[0,1]$. Since $\mathcal F$ is linear, the Bayesian inverse problem can be solved exactly using \eqref{eq:mapMuAffine} and \eqref{eq:mapGammaAffine}, without the need for numerical methods. Therefore, this toy problem serves solely as a numerical validation of \Cref{thm:convergence}, rather than an alternative to computing the exact solution. Given the linearity of $\mathcal F$, we employ a variational autoencoder with the architecture shown in \Cref{fig:linearVAE} for both the eUQ-VAE and UQ-VAE. The encoder has no hidden layers and uses linear activation functions.\\
We generate a single sample $(\mathbf u^\mathrm{(1)}, \tilde{ \mathbf y}^\mathrm{(1)})$, where $\mathbf u^\mathrm{(1)}$ is sampled from $\mathcal N(\boldsymbol \mu_\mathrm{pr},\Gamma_\mathrm{pr})$ and $ \tilde{ \mathbf y}^\mathrm{(1)} = F \mathbf u^\mathrm{(1)}$. We assume $\boldsymbol \mu_\mathrm{E} = \mathbf 0$ and $\Gamma_\mathrm{E}$ diagonal with $(\Gamma_\mathrm{E})_\mathrm{i,i} = \eta^\mathrm{2} \tilde{y}^\mathrm{(1)2}_\mathrm{i}$ for $i = 1,\dots, O$, where $\eta \in \R$. We sample $\boldsymbol \epsilon^\mathrm{(1)} \sim \mathcal N (\boldsymbol \mu_\mathrm{E}, \Gamma_\mathrm{E})$ and add noise to the observational data $\mathbf y^\mathrm{(1)} = \tilde{\mathbf y}^\mathrm{(1)}+\boldsymbol \epsilon^\mathrm{(1)}$. The corresponding maximum a posteriori parameter and Laplace approximation of the covariance matrix are denoted as  $( \mathbf u_\mathrm{MAP}^\mathrm{(1)},\Gamma_\mathrm{Lap}^\mathrm{(1)} )$.

\begin{figure}[t!]
	\centering
 	\includegraphics[width=\linewidth]{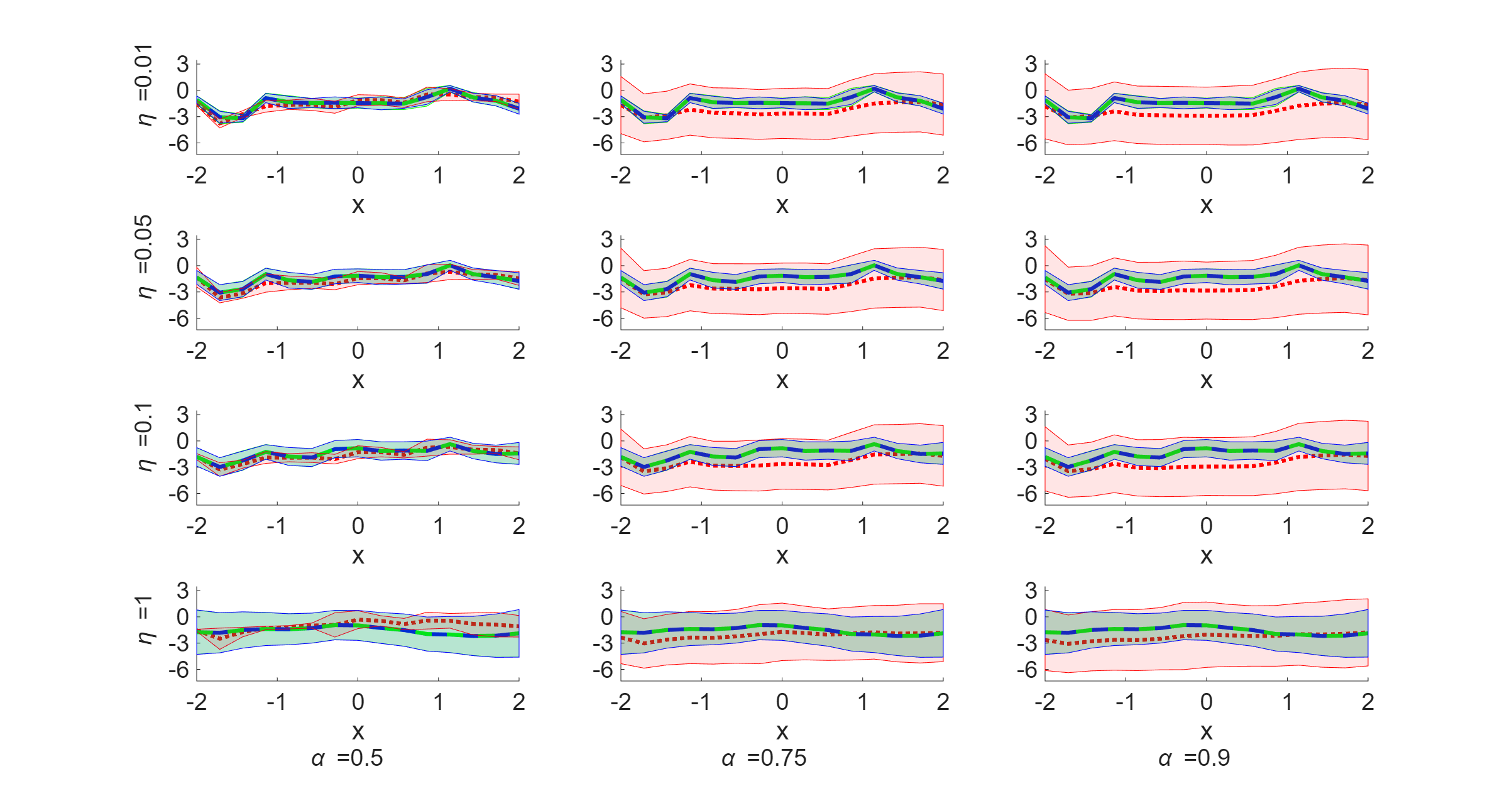}
	\caption{True value of $\mathbf u_\mathrm{MAP}^\mathrm{(1)}$ and its $\mathrm{tr} (\Gamma_\mathrm{Lap}^\mathrm{(1)} )$ (green dashed lines and areas) and their estimates by means of eUQ-VAE (blue dashed lines and areas) and UQ-VAE (red dotted lines and areas). According to the rows and columns, we modify the values of $\eta$ and $\alpha$, respectively. The green and blue lines and areas overlap as expected by \Cref{cor:convergenceNN}.}
	\label{fig:test1}
\end{figure}

In \Cref{fig:test1}, we compare the posterior mean estimates and their standard deviations, obtained using the eUQ-VAE and the UQ-VAE, for different values of $\eta$ and $\alpha$. For reference, we report the loss function for the UQ-VAE, where a single-sample Monte Carlo is used to estimate the expected values $\E_\mathrm{p_\mathrm{U}(\mathbf u)} \left[ \norm{\boldsymbol \mu-\mathbf u}_\mathrm{\Gamma^\mathrm{-1}}^\mathrm{2} \right]$ and $\E_\mathrm{q_\mathrm{\phi}(\mathbf u | \mathbf y)}\left[ \norm{\mathbf y - \boldsymbol \mu_\mathrm{E} - \mathcal F(\mathbf u)}_\mathrm{\Gamma_\mathrm{E}^\mathrm{-1}}^\mathrm{2} \right]$ in place of $\E_\mathrm{q_\mathrm{pro,\phi}(\mathbf u | \mathbf y)}\left[ \norm{\mathbf y - \boldsymbol \mu_\mathrm{E} - \mathcal F(\mathbf u)}_\mathrm{\Gamma_\mathrm{E}^\mathrm{-1}}^\mathrm{2} \right]$, appearing in the fifth row of \eqref{eq:pUexpval} and in \eqref{eq:qUexpval}, respectively. The loss function for UQ-VAE is given by:
\begin{align}
 \tilde {L}_\mathrm{\alpha}(\boldsymbol \mu,\Gamma;\mathcal F) =  &\frac{(1-\alpha)}{\alpha} \left( \log |\Gamma| + \norm{\boldsymbol \mu - \mathbf u^\mathrm{(1)}}_\mathrm{\Gamma^\mathrm{-1}}^\mathrm{2} \right)\\
& + \norm{\mathbf y - \boldsymbol \mu_\mathrm{E} - \mathcal F(\mathbf u^\mathrm{1})}_\mathrm{\Gamma_\mathrm{E}^\mathrm{-1}}^\mathrm{2} \\
 &- \log |\Gamma| + \norm{\boldsymbol \mu - \boldsymbol \mu_\mathrm{pr}}_\mathrm{\Gamma_\mathrm{pr}^\mathrm{-1}}^\mathrm{2} +\mathrm{tr}\left( \Gamma_\mathrm{pr}^\mathrm{-1} \Gamma  \right), \label{eq:gohProblem}
\end{align}
where $\mathbf u^\mathrm{1}$ is sampled from $\mathcal N(\boldsymbol \mu,\Gamma)$, $\mathbf u^\mathrm{(1)}$ is the true parameter and $(\boldsymbol \mu_\mathrm{post},\Gamma_\mathrm{post})=(\hat{\boldsymbol \mu}, \hat{\Gamma})$, where $(\hat{\boldsymbol \mu}, \hat{\Gamma})$ denote the stationary points of $\tilde {L}_\mathrm{\alpha}(\boldsymbol \mu,\Gamma;\mathcal F)$.\\
The eUQ-VAE outperforms the UQ-VAE in terms of accuracy. It is robust with respect to both $\eta$ and $\alpha$. In accordance of $( \mathbf u_\mathrm{MAP}^\mathrm{(1)},\Gamma_\mathrm{Lap}^\mathrm{(1)} )$ being independent of $\alpha$, the eUQ-VAE estimates remain constant as $\alpha$ varies. In contrast, the UQ-VAE estimates lack robustness with respect to $\alpha$, leading to an overestimation of the standard deviation for large values of $\alpha$ and inaccurate mean estimates. This comes at an increased computational cost to evaluate the loss function \eqref{eq:minProblem3}. Its computational complexity is $O(D^\mathrm{3}+KD^\mathrm{2}+K\mathrm{c}(\mathcal F) + KO^\mathrm{2})$ (where $c(\mathcal F)$ is the cost of a single evaluation of $\mathcal F$), compared to $O(D^\mathrm{3}+O^\mathrm{2} +\mathrm{c}(\mathcal F))$ for the loss function \eqref{eq:gohProblem}. Specifically, in \eqref{eq:minProblem3}, the term $D^\mathrm{2}$ corresponds to computing a single $\mathbf u^\mathrm{k}$, $\mathrm{c}(\mathcal F)$ to evaluating $\mathcal F$ and $O^\mathrm{2}$ to solving linear systems involving $\Gamma_\mathrm{E}$ (assuming its Cholesky factorization is known). These operations are repeated $K$ times. The $D^\mathrm{3}$ term accounts for solving the linear systems $\Gamma^\mathrm{-1}\Gamma_\mathrm{pr}$ and $\Gamma_\mathrm{pr}^\mathrm{-1}
\Gamma$. The cost of all these operations dominates that of the other ones in \eqref{eq:minProblem3}.

\subsection{Training of eUQ-VAE over a large dataset} \label{sec:test2}
We now assess the generalization properties of the eUQ-VAE and compare them with those of UQ-VAE. The distribution of $U$, the function $\mathcal F$, $D$, $O$ and the VAE architecture are the same as in \Cref{sec:test1}. Additionally, we generate the training dataset $\{(\mathbf u^\mathrm{(m)}, \tilde{ \mathbf y}^\mathrm{(m)})\}_\mathrm{m=1}^\mathrm{M}$, where $\mathbf u^\mathrm{(m)}$ is drawn from $\mathcal N(\boldsymbol \mu_\mathrm{pr},\Gamma_\mathrm{pr})$, $ \tilde{ \mathbf y}^\mathrm{(m)} = F \mathbf u^\mathrm{(m)}$ and $M\in \N$. We assume $\boldsymbol \mu_\mathrm{E} = \mathbf 0$ and $\Gamma_\mathrm{E}$ is diagonal, with $(\Gamma_\mathrm{E})_\mathrm{i,i} = \eta^\mathrm{2} \underset{m}{\max} \, \abs{\tilde{y}^\mathrm{(m)}_\mathrm{i}}^\mathrm{2}$ for $i = 1,\dots, O$, where $\eta \in \R$. We sample the noise $\boldsymbol \epsilon^\mathrm{(m)} \sim \mathcal N (\boldsymbol \mu_\mathrm{E}, \Gamma_\mathrm{E})$ and add it to the observational data $\mathbf y^\mathrm{(m)} = \tilde{\mathbf y}^\mathrm{(m)}+\boldsymbol \epsilon^\mathrm{(m)}$ for $m = 1,\dots,M$. The corresponding maximum a posteriori estimates and Laplace approximations of the covariance matrix are denoted as $\{ ( \mathbf u_\mathrm{MAP}^\mathrm{(m)},\Gamma_\mathrm{Lap}^\mathrm{(m)} ) \}_\mathrm{m=1}^\mathrm{M}$. We generate the test set analogously, with a size equal to $1/8$ of the training set.

\begin{figure}[t!]
	\centering
 	\includegraphics[width=\linewidth]{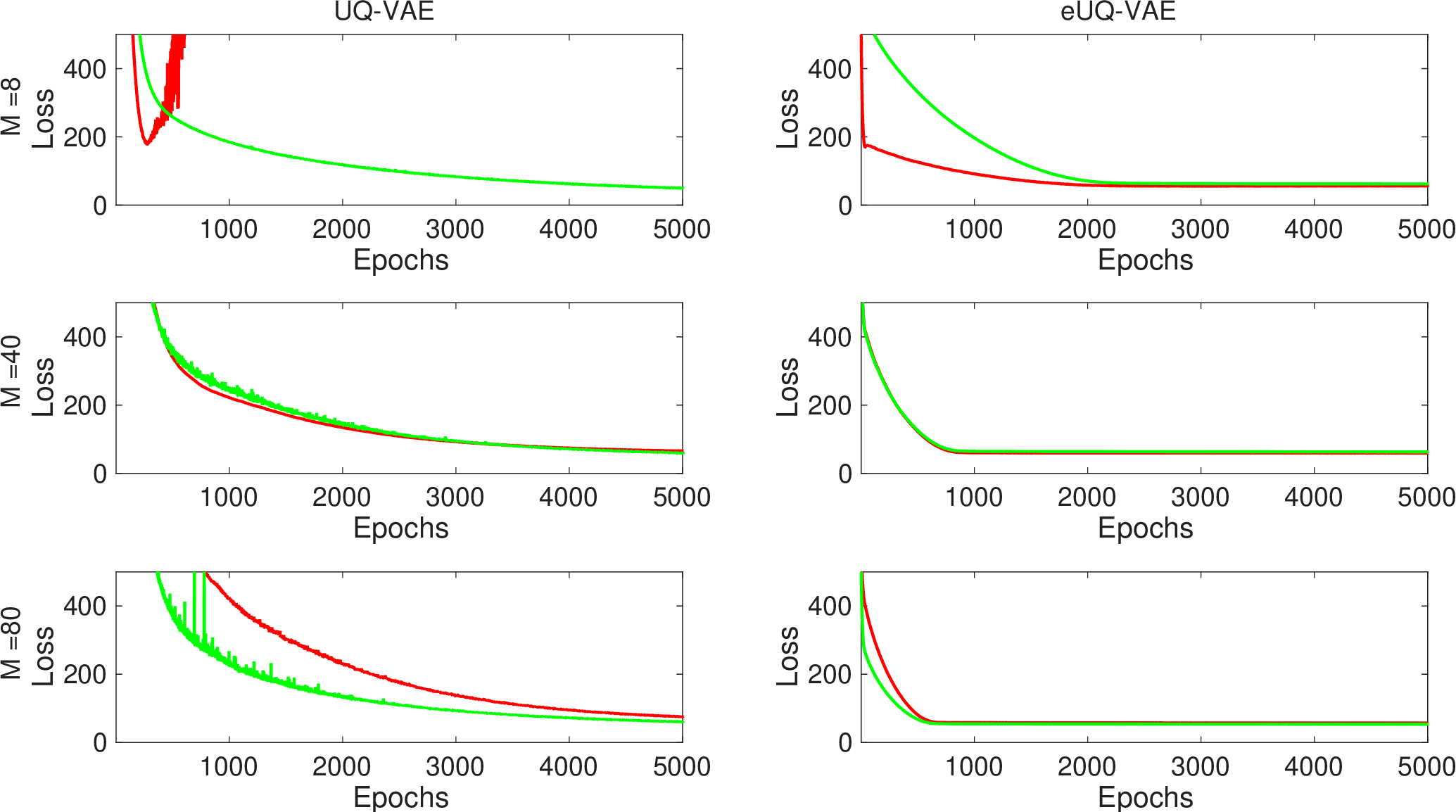}
	\caption{Training (light green) and test (red) loss values for UQ-VAE and eUQ-VAE with $\eta = 0.05$ and $\alpha = 0.5$. Each row corresponds to a different training set size $M$.}
	\label{fig:test2}
\end{figure}

We compare the generalization capabilities of eUQ-VAE and UQ-VAE across different training set sizes $M$, with $\eta = 0.05$ and $\alpha = 0.5$. We train the VAEs for $5000$ epochs (\Cref{fig:test2}). Although this exceeds the number of epochs strictly necessary, eUQ-VAE converges faster than UQ-VAE and shows no signs of overfitting, even with small training sets. Beyond convergence speed, eUQ-VAE maintains consistent performance across all considered dataset sizes. In contrast, UQ-VAE converges more slowly and overfits on small datasets, though this effect diminishes as the dataset size grows. Note that direct comparison of loss values between UQ-VAE and eUQ-VAE is not meaningful, as the two approaches employ different loss functions.

\begin{figure}[t!]
	\centering
 	\includegraphics[width=\linewidth]{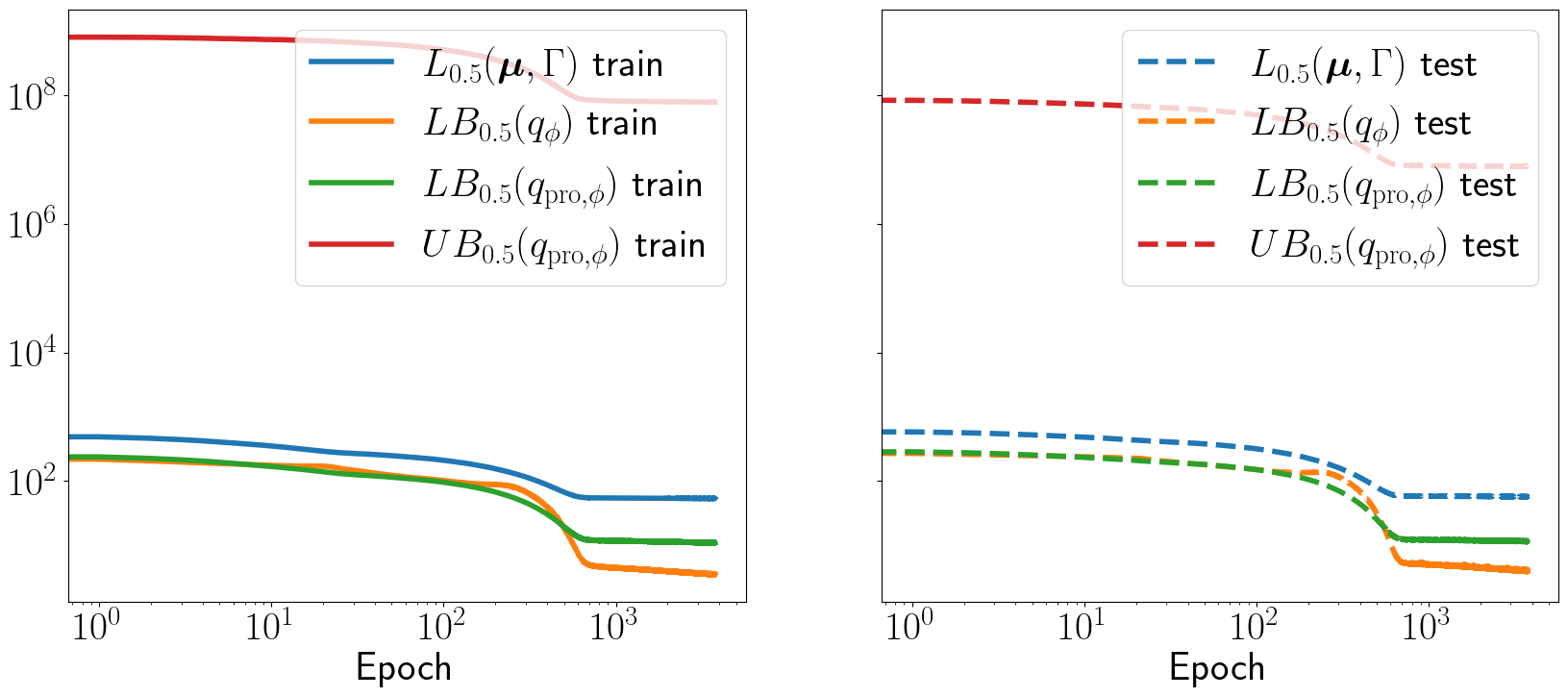}
	\caption{Values of the loss function \eqref{eq:minProblem3}, lower bounds \eqref{eq:lowerBound} and upper bounds \eqref{eq:minProblem} on the training and test datasets obtained with the eUQ-VAE approach for $\eta = 0.05$ and $\alpha = 0.5$.}
	\label{fig:test2_losses}
\end{figure}

We analyze the evolution of several quantities during eUQ-VAE training on both the training and test datasets (\Cref{fig:test2_losses}). The training set consists of $80$ samples, while the test one of $10$ samples. An early stopping criterion with a patience of $100$ epochs is used to halt the training once the loss function minimum is reached. For $\alpha = 0.5$, it holds $LB_\mathrm{0.5}(q_\mathrm{pro,\phi}) \le UB_\mathrm{0.5}(q_\mathrm{pro,\phi}) \sim L_\mathrm{0.5}(\boldsymbol \mu, \Gamma)$, where $\sim$ denotes equality up to constants in $(\boldsymbol \mu, \Gamma)$.\\
On both training and test sets, $L_\mathrm{0.5}(\boldsymbol \mu, \Gamma)$ and $UB_\mathrm{0.5}(q_\mathrm{pro,\phi})$ exhibit the same trends, as the upper bound is essentially a scaled and shifted version of the loss function.\\
In the affine case, the lower bounds $LB_\mathrm{\alpha}(q_\mathrm{pro,\phi})$ and $LB_\mathrm{\alpha}(q_\mathrm{\phi})$, which quantify the discrepancy between the input distribution and the posterior one $p_\mathrm{U|Y}(\mathbf u| \mathbf y)$, are explicitly computable. Specifically, if $q \sim \mathcal N(\check \mu, \check \Gamma)$ is independent of $U|Y$, the term $(1-\alpha)q+\alpha p$ from the JSD family \eqref{eq:JSDfamily} is Gaussian, with distribution $\mathcal N((1-\alpha)\check \mu+\alpha \mathbf u_\mathrm{MAP},(1-\alpha)^\mathrm{2} \check \Gamma+\alpha^\mathrm{2} \Gamma_\mathrm{Lap})$. The KL divergence between two Gaussian distributions $\mathcal N_\mathrm{0}(\mu_\mathrm{0},\Gamma_\mathrm{0})$ and $\mathcal N_\mathrm{1}(\mu_\mathrm{1},\Gamma_\mathrm{1})$ is given by:
\begin{align}
KL(\mathcal N_\mathrm{0}|| \mathcal N_\mathrm{1}) =  \frac{1}{2}\left(\mathrm{tr}(\Gamma_\mathrm{1}^\mathrm{-1}\Gamma_\mathrm{0}) + \norm{\boldsymbol \mu_\mathrm{1}-\boldsymbol \mu_\mathrm{0}}_\mathrm{\Gamma_\mathrm{1}^\mathrm{-1}}^\mathrm{2}-D+\log \frac{|\Gamma_\mathrm{1}|}{|\Gamma_\mathrm{0}|} \right).
\end{align}
Since, in the affine case, all terms in $LB_\mathrm{\alpha}$ are KL divergences between Gaussian distributions, the lower bound can be  computed explicitly. Initially, the two lower bounds are nearly identical due to eUQ-VAE weights initialization (\Cref{sec:DN_WI}) for $\alpha=0.5$. Rewriting \eqref{eq:muMap2} and \eqref{eq:gammaMap2}, at the first iteration we have:
\begin{align}
\boldsymbol \mu_\mathrm{post}^\mathrm{(m)} &= \frac{1-\alpha}{\alpha} \Gamma_\mathrm{Lap}^\mathrm{(m)} \Gamma^\mathrm{-1} (\boldsymbol \mu^\mathrm{(m)}-\boldsymbol \mu_\mathrm{pr}) + \boldsymbol \mu^\mathrm{(m)}\\
&\sim \frac{1-\alpha}{\alpha} \Gamma_\mathrm{Lap}^\mathrm{(m)} \Gamma^\mathrm{-1}_\mathrm{pr} (\boldsymbol \mu_\mathrm{pr}-\boldsymbol \mu_\mathrm{pr}) + \boldsymbol \mu_\mathrm{pr} =\boldsymbol \mu_\mathrm{pr} \sim \boldsymbol \mu^\mathrm{(m)},\\
\Gamma_\mathrm{post}^\mathrm{(m)} &= \frac{\alpha}{1-\alpha} \Gamma \left((\boldsymbol \mu^\mathrm{(m)}- \boldsymbol \mu_\mathrm{pr})(\boldsymbol \mu^\mathrm{(m)}- \boldsymbol \mu_\mathrm{pr})^\mathrm{T} + \Gamma_\mathrm{pr} \right)^\mathrm{-1} \Gamma^\mathrm{(m)} \\
&\sim  \Gamma_\mathrm{pr} \left((\boldsymbol \mu_\mathrm{pr}- \boldsymbol \mu_\mathrm{pr})(\boldsymbol \mu_\mathrm{pr}- \boldsymbol \mu_\mathrm{pr})^\mathrm{T} + \Gamma_\mathrm{pr} \right)^\mathrm{-1} \Gamma_\mathrm{pr} = \Gamma_\mathrm{pr} \sim \Gamma^\mathrm{(m)}.
\end{align}
Towards the end of training, the lower bound computed in $q_\mathrm{\phi}(\mathbf u | \mathbf y)$ becomes smaller than that in $q_\mathrm{pro,\phi}(\mathbf u | \mathbf y)$, indicating that the transformation of the eUQ-VAE latent states improves the approximation of the posterior distribution.

\subsection{Laplace problem}\label{sec:test3}
We consider the Laplace equation on the unit square $\Omega = (0,1)^\mathrm{2}$:
\begin{align}
\begin{cases}
-\nabla \cdot(e^\mathrm{u}\nabla y) = 0 \qquad & \text{in } \Omega,\\
-e^\mathrm{u}(\nabla y \cdot \mathbf n) = -1 \qquad & \text{on } \Omega^\mathrm{root} \label{eq:heatConduction},\\
-e^\mathrm{u}(\nabla y \cdot \mathbf n) = \frac{y}{2} \qquad & \text{on } \partial \Omega\setminus \Omega^\mathrm{root},
\end{cases}
\end{align}
where $u$ is the (log) diffusion parameter, $\mathbf n$ denotes the outward unit normal vector to the domain and $\Omega^\mathrm{root} =  [0,1] \times \{0\}$.

The parameters $U$ are distributed as a Gaussian random field $\mathcal N (\mu_\mathrm{pr},\mathcal C_\mathrm{pr})$ on $\Omega$, with prior mean $ \mu_\mathrm{pr}= 0$. To compute the covariance operator  $\mathcal C_\mathrm{pr}$, we define the differential operator $\mathcal A$ as:
\begin{align}
\mathcal A u = -\gamma \Delta u + \delta u \qquad \text{in } \Omega,
\end{align}
where $\gamma>0, \delta >0$. The covariance operator $\mathcal C_\mathrm{pr}$ is defined as $G_\mathrm{2}$, the Green function solution of:
\begin{align}
\begin{cases}
\mathcal A G_\mathrm{1} = \delta_\mathrm{\mathbf x} \qquad &\text{in } \Omega,\\
\nabla G_\mathrm{1} \cdot \mathbf n + \beta G_\mathrm{1} = 0  &\text{on } \partial \Omega,\\
\mathcal A G_\mathrm{2} = G_\mathrm{1} & \text{in } \Omega,\\
\nabla G_\mathrm{2} \cdot \mathbf n + \beta G_\mathrm{2} = 0 & \text{on } \partial \Omega,
\end{cases} 
\end{align}
where, for $\mathbf x \in \Omega$, $\delta_\mathrm{\mathbf x}$ denotes the Dirac delta centered in $\mathbf x$ and $\beta$ is chosen to mitigate the boundary artifacts in the Green function \cite{daon2018mitigating}. We set $\gamma = 0.1$ and $\delta = 0.5$. Priors of this form guarantee the well-posedness of infinite dimensional Bayesian inverse problems\cite{stuart2010inverse,bui2013computational}. To compute $\mathcal C_\mathrm{pr}$, we employ the Python library for inverse problems \emph{hIPPYlib} \cite{VillaPetraGhattas21,VillaPetraGhattas18,VillaPetraGhattas16}, using linear finite element methods on a triangular mesh of $\Omega$ with $D = 17^\mathrm{2}$ degrees of freedom. The same mesh is used to solve \eqref{eq:heatConduction}. We define $(\boldsymbol{\mu}_\mathrm{pr})_\mathrm{i} = \mu_\mathrm{pr}(\mathbf x_\mathrm{i})$ and $(\Gamma_\mathrm{pr})_\mathrm{i,j} = \mathcal C_\mathrm{pr}(\mathbf x_\mathrm{i},\mathbf x_\mathrm{j})$ for $i,j = 1,\cdots,D$, where $\mathbf x_\mathrm{i}$ are the coordinates of the degrees of freedom.\\
The observational data consist of $O=20$ randomly distributed evaluations of the solution to \eqref{eq:heatConduction}.

We generate the training dataset $\{(\mathbf u^\mathrm{(m)}, \tilde{ \mathbf y}^\mathrm{(m)})\}_\mathrm{m=1}^\mathrm{M}$, sampling $\mathbf u^\mathrm{(m)}$ from $\mathcal N(\boldsymbol \mu_\mathrm{pr},\Gamma_\mathrm{pr})$, $ \tilde{ \mathbf y}^\mathrm{(m)}$ are the observations of the solution to \eqref{eq:heatConduction} with diffusive parameter $\mathbf u^\mathrm{(m)}$ and $M \in \mathbb N$. We assume $\boldsymbol \mu_\mathrm{E} = \mathbf 0$ and $\Gamma_\mathrm{E}$ is diagonal with $(\Gamma_\mathrm{E})_\mathrm{i,i} = \eta^\mathrm{2} \underset{j,m}{\max} \, \abs{\tilde{y}^\mathrm{(m)}_\mathrm{j}}^\mathrm{2}$ for $i = 1,\dots, O$, where $\eta \in \R$. We sample $\boldsymbol \epsilon^\mathrm{(m)} \sim \mathcal N (\boldsymbol \mu_\mathrm{E}, \Gamma_\mathrm{E})$ and perturb the observational data $\mathbf y^\mathrm{(m)} = \tilde{\mathbf y}^\mathrm{(m)}+\boldsymbol \epsilon^\mathrm{(m)}$ for $m=1,\cdots,M$. We fix $\eta = 0.05$ and $\alpha = 0.5$.

We use loss function \eqref{eq:minProblem4} to train the encoder. The decoder $\psi$ is trained prior to the encoder using the root mean squared error loss. We train the decoder on a dataset of $M=1600$ samples to accurately approximate the map $\mathcal F$, and subsequently train the encoder on a dataset of $80$ samples. The decoder architecture consists of six hidden layers with $100$ neurons each, achieving a final validation loss of order $10^\mathrm{-2}$. The encoder comprises two hidden layers each with $1000$ neurons.

The ground-truth $\{ (\mathbf u_\mathrm{MAP}^\mathrm{(m)}, \Gamma_\mathrm{Lap}^\mathrm{(m)}) \}$, for $m=1,\dots,M$, are computed using the Python library \emph{hIPPYlib}.

\begin{figure}[t!]
	\centering
	\subfigure[][From left to right: solution and true parameter of the Laplace equation. For the solution the observational points are highlighted. Posterior parameter estimates with eUQ-VAE and MAP approaches. \label{fig:1samplea}]{\includegraphics[width=\linewidth]{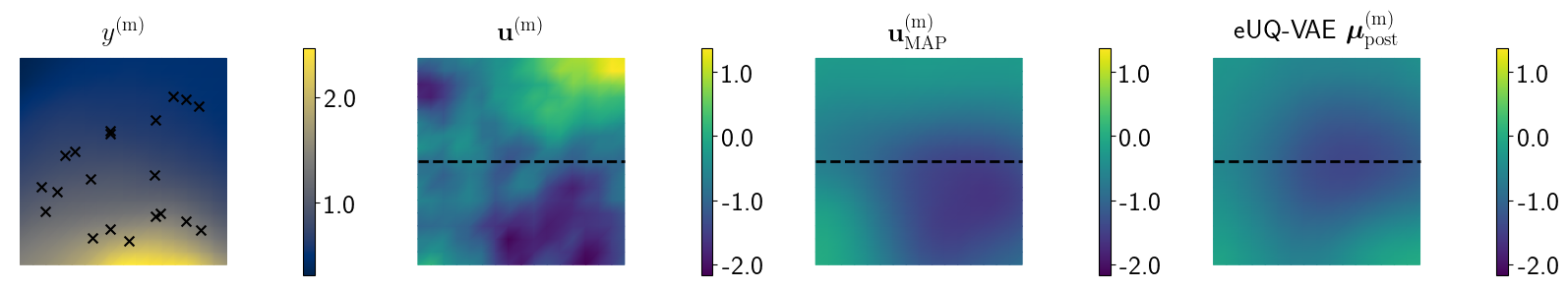}} \\
	\subfigure[][Posterior variance estimates. \label{fig:1sampleb}]{\includegraphics[width=0.495\linewidth]{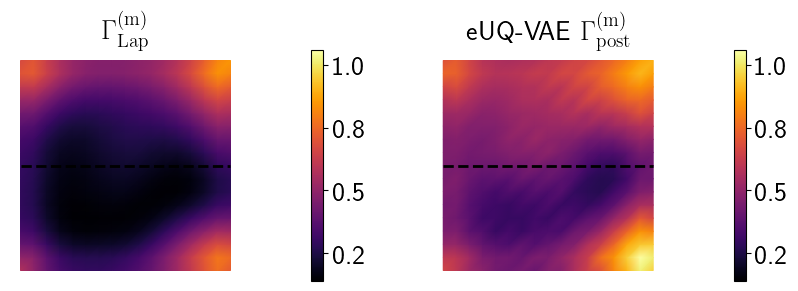}}
	\hfill
 	\subfigure[][Estimated parameter and standard deviation for $y=0.5$. \label{fig:1samplec}]{\includegraphics[width=0.495\linewidth]{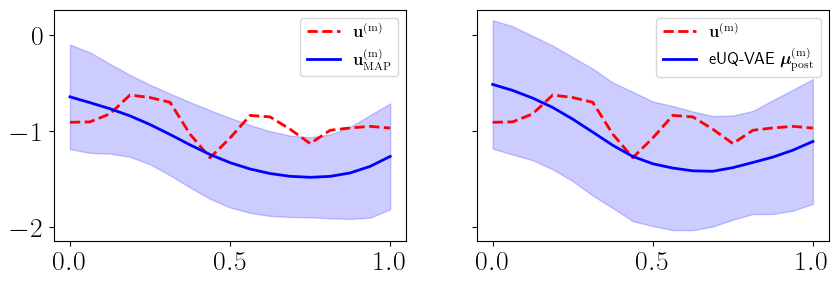}}
\caption{Test sample 1.}\label{fig:1sample}
\end{figure}

\begin{figure}[t!]
	\centering
 	\subfigure[][From left to right: solution and true parameter of the Laplace equation. For the solution the observational points are highlighted. Posterior parameter estimates with eUQ-VAE and MAP approaches. \label{fig:2samplea}]{\includegraphics[width=\linewidth]{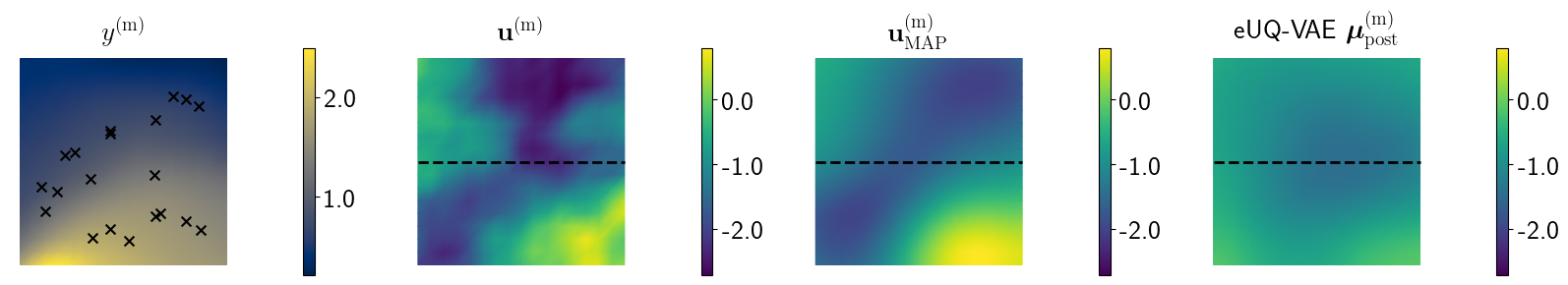}} \\
	\subfigure[][Posterior variance estimates. \label{fig:2sampleb}]{\includegraphics[width=0.495\linewidth]{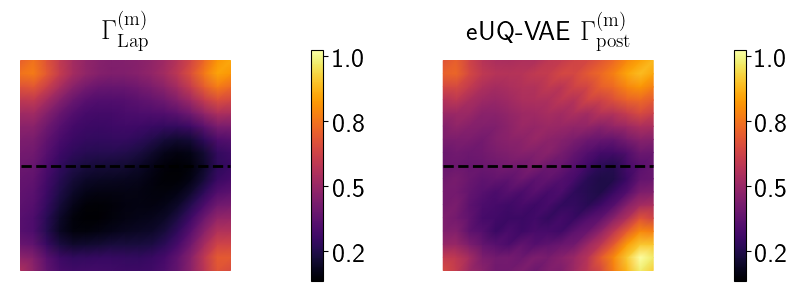}}
	\hfill
 	\subfigure[][Estimated parameter and standard deviation for $y=0.5$. \label{fig:2samplec}]{\includegraphics[width=0.495\linewidth]{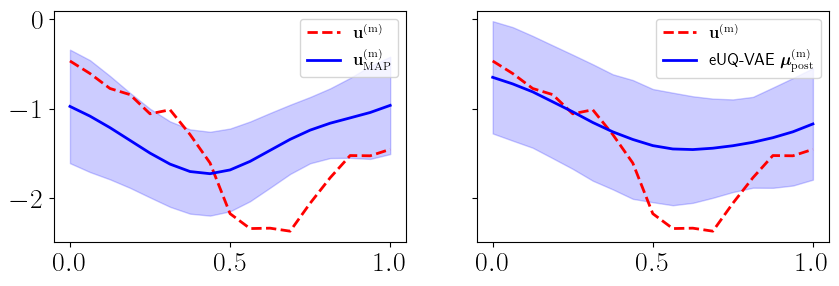}}
	\caption{Test sample 2.} \label{fig:2sample}
\end{figure}

\begin{figure}[t!]
	\centering
 	\subfigure[][From left to right: solution and true parameter of the Laplace equation. For the solution the observational points are highlighted. Posterior parameter estimates with eUQ-VAE and MAP approaches. \label{fig:3samplea}]{\includegraphics[width=\linewidth]{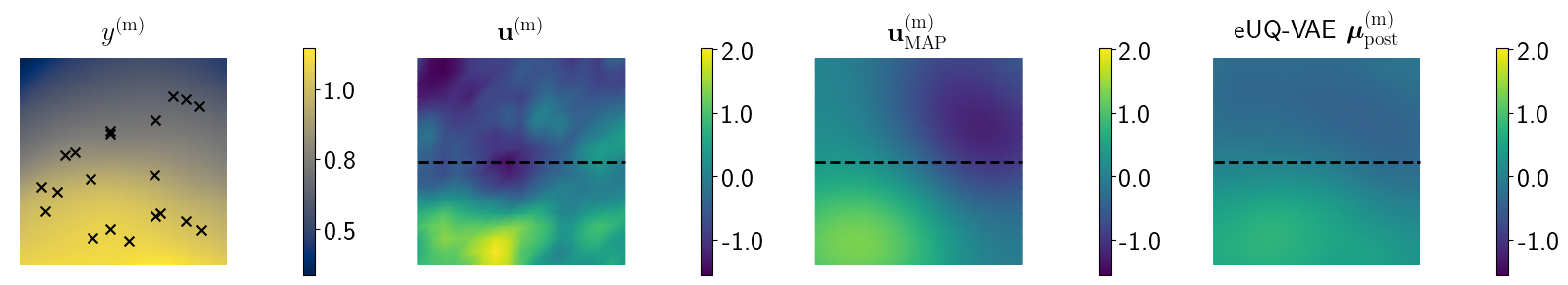}} \\
	\subfigure[][Posterior variance estimates. \label{fig:3sampleb}]{\includegraphics[width=0.495\linewidth]{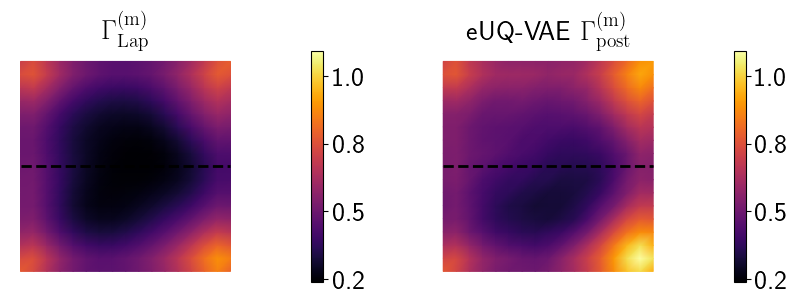}}
	\hfill
 	\subfigure[][Estimated parameter and standard deviation for $y=0.5$. \label{fig:3samplec}]{\includegraphics[width=0.495\linewidth]{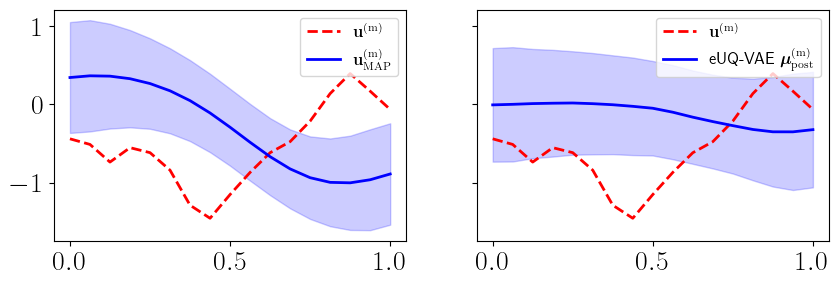}}
	\caption{Test sample 3.} \label{fig:3sample}
\end{figure}

We evaluate the eUQ-VAE on three test samples. The estimated means capture the trends of $\mathbf u_\mathrm{MAP}^\mathrm{(m)}$ (\Cref{fig:1samplea,fig:1samplec,fig:2samplea,fig:2samplec,fig:3samplea,fig:3samplec}), even if they fail to reproduce bigger oscillations in $\mathbf u_\mathrm{MAP}^\mathrm{(m)}$. Since $\mathbf u_\mathrm{MAP}^\mathrm{(m)}$ represents the most likely parameter generating the available data and that the eUQ-VAE parameter is an approximation of it, discrepancies between the estimates $\mathbf u_\mathrm{MAP}^\mathrm{(m)}$ and $\boldsymbol \mu_\mathrm{post}^\mathrm{(m)}$ and the true parameter $\mathbf u^\mathrm{(m)}$ may occur. The variance $\mathrm{tr}(\Gamma_\mathrm{post}^\mathrm{(m)})$ estimated by the eUQ-VAE captures the scale of the Laplace approximation $\mathrm{tr}(\Gamma_\mathrm{Lap}^\mathrm{(m)})$ (\Cref{fig:1sampleb,fig:1samplec,fig:2sampleb,fig:2samplec,fig:3sampleb,fig:3samplec}), though it tends to overestimate it.

We compare the computational cost and accuracy of eUQ-VAE with those of UQ-VAE approach, where the encoder and the decoder are trained simultaneously using \eqref{eq:gohProblem}, and with MCMC based on the Metropolis-Hastings algorithm \cite{hastings1970monte} with $10^\mathrm{4}$ samples.\\
Because eUQ-VAE requires to approximate an expectation with a large number of samples ($2^\mathrm{12}$), its training time is several times longer than that of UQ-VAE (\Cref{table:times}). On average, the estimation of $\mathbf u_\mathrm{MAP}$ takes $0.17$s per sample with eUQ-VAE, compared to $0.06$s with UQ-VAE. The difference arises because $(\boldsymbol \mu_\mathrm{post}^\mathrm{(m)},\Gamma_\mathrm{post}^\mathrm{(m)})$ are computed as a linear transformation of the encoder outputs in eUQ-VAE, while in UQ-VAE they coincide with the encoder outputs. Both methods outperform MCMC in terms of computational time, which requires on average $79$s to solve a Bayesian inverse problem.
\begin{table}[t!]
\footnotesize
\centering
\begin{tabular}{|c|c|c|}
 \hline
 & Offline phase & Online phase (100 samples)\\
 \hline
 eUQ-VAE & $50.9$h & $17$s\\
 UQ-VAE & $0.4$h & $6$s\\
 MCMC & - & $2.2$h\\
 \hline 
\end{tabular}
\label{table:times}
\caption{Training times (offline phase) of the UQ-VAE and eUQ-VAE and solution times of $100$ Bayesian inverse problems for eUQ-VAE, UQ-VAE and MCMC.}
\end{table}

\begin{figure}[t!]
\centering
\includegraphics[width=\linewidth]{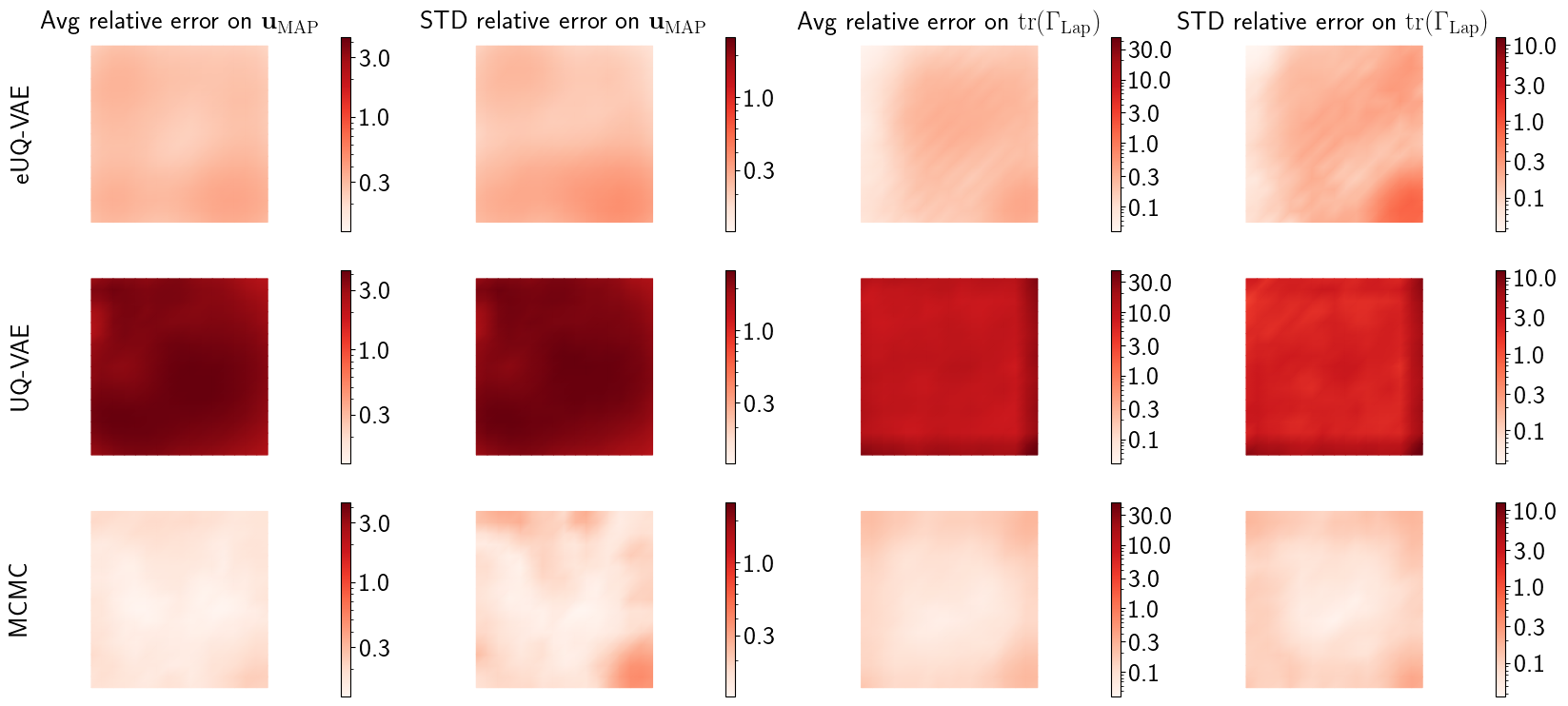}
\caption{Average errors and standard deviations of eUQ-VAE, UQ-VAE and MCMC at estimating $\mathbf u_\mathrm{MAP}^\mathrm{(m)}$ and $\Gamma_\mathrm{Lap}^\mathrm{(m)}$ on a test dataset of $100$ samples. Results are shown on a logarithmic scale to facilitate comparisons across methods.}
\label{fig:errors}
\end{figure}

We assess accuracy by computing the relative error in the maximum a posteriori parameter $\mathbf u_\mathrm{MAP}^\mathrm{(m)}$ and in the Laplace approximation of the variances $\mathrm {tr}( \Gamma_\mathrm{Lap}^\mathrm{(m)})$ of the $m$-th sample, measured in the infinity norm at each degree of freedom:
\begin{gather}
\mathbf {e}_\mathrm{\mathbf u_\mathrm{MAP}^\mathrm{(m)}} = \frac{\abs{\boldsymbol \mu_\mathrm{post}^\mathrm{(m)}-\mathbf u_\mathrm{MAP}^\mathrm{(m)}}}{\norm{\mathbf u_\mathrm{MAP}^\mathrm{(m)}}_\mathrm{\infty}}, \qquad
\mathbf {e}_\mathrm{\mathrm {tr}( \Gamma_\mathrm{Lap}^\mathrm{(m)})} = \frac{\abs{\mathrm{diag}(\Gamma_\mathrm{post}^\mathrm{(m)})-\mathrm {tr}( \Gamma_\mathrm{Lap}^\mathrm{(m)})}}{\norm{\mathrm {tr}( \Gamma_\mathrm{Lap}^\mathrm{(m)})}_\mathrm{\infty}}.
\end{gather}
Here $(\boldsymbol \mu_\mathrm{post}^\mathrm{(m)},\Gamma_\mathrm{post}^\mathrm{(m)})$ are the UQ-VAE estimates for the $m$-th sample, $\abs{\cdot}$ is applied element-wise and $\norm{\cdot}_\mathrm{\infty}$ denotes the vector infinite norm. The maximum of these errors over the mesh provides the relative error on the $m$-th sample in the infinity norm.\\
Although MCMC is the slowest method during the online phase, it yields the most accurate estimations (\Cref{fig:errors}). The eUQ-VAE produces slightly larger relative errors in approximating both $\mathbf u_\mathrm{MAP}^\mathrm{(m)}$ and $\mathrm{tr}(\Gamma_\mathrm{Lap}^\mathrm{(m)})$, but its accuracy is consistent across samples, as shown by the low standard deviations of the errors. In contrast, UQ-VAE performs poorly in estimating both $\mathbf u_\mathrm{MAP}^\mathrm{(m)}$ and $\Gamma_\mathrm{Lap}^\mathrm{(m)}$.

\section{Conclusions}\label{sec:concl}
In this work, we introduced an enhanced uncertainty quantification variational autoencoder (eUQ-VAE) for solving Bayesian inverse problems, building on the work of  \cite{goh2021solving}. We proved, in the affine case, a theorem showing that the stationary points of the novel loss function proposed for training the eUQ-VAE are connected to the posterior distribution of the parameters. This theoretical result is established for single-layer linear neural networks. In \Cref{sec:test1}, we showed (in the affine case) that the eUQ-VAE estimates the posterior mean and covariance more accurately than the UQ-VAE. Moreover, the eUQ-VAE exhibits superior generalization properties on small datasets and converges in fewer epochs compared to the UQ-VAE (\Cref{sec:test2}). For a Laplace equation, the posterior mean estimated by eUQ-VAE generally aligns with the MAP estimate (\Cref{sec:test3}. however, the eUQ-VAE tends to slightly overestimate the variance.

The eUQ-VAE enables real-time Bayesian inverse problem solution and uncertainty quantification after an (expensive) offline training phase. Additionally, the training is not subjected to the knowledge of the posterior distribution mean and covariance that the eUQ-VAE aims to estimate. This new methodology can be applied to a wide range of inverse problems, as the VAE loss function is independent of the specific application.

Despite the advantages, some limitations remain. The estimation of $\E_\mathrm{q_\mathrm{pro,\phi}(\mathbf u | \mathbf y)}\left[ \norm{\mathbf y - \boldsymbol \mu_\mathrm{E} - \mathcal F(\mathbf u)}_\mathrm{\Gamma_\mathrm{E}^\mathrm{-1}}^\mathrm{2} \right]$ within the eUQ-VAE training is the main contributor to its high training cost. In high dimensional spaces, the number of samples required to approximate this expectation increases due to the curse of dimensionality. A different treatment of this term could remove the need for estimation, thereby reducing training time while improving accuracy.\\
\Cref{thm:convergence} currently states that the loss function is separately convex in the mean and covariance of the proxy distribution. Proving joint convexity would strengthen the result by guaranteeing that minimizing the loss leads to the estimation of the posterior distribution. Moreover, the convergence analysis is limited to the affine case and an extension to general parameters to outputs maps would improve the mathematical foundations of the eUQ-VAE.\\
Finally, the decoder NN is trained using both parameters and observational data, whereas the encoder NN requires only the observational data. The prior distribution of the parameters suffices for training the encoder and estimating the posterior. Developing a decoder NN that can be trained without parameters, would enhance the applicability of the eUQ-VAE, as in many real case scenarios only observational data are available.

\section*{Acknowledgments}
AT and LD are members of the INdAM group GNCS \say{Gruppo Nazionale per il Calcolo Scientifico} (National Group for Scientific Computing). 

AT and LD acknowledge the INdAM GNCS project CUP E53CE53C24001950001.

LD acknowledges the support by the FAIR (\say{Future Artificial Intelligence Research}) project, funded by the NextGenerationEU program within the PNRR-PE-AI scheme (M4C2, investment 1.3, line on Artificial Intelligence), Italy.

LD acknowledges the project PRIN2022, MUR, Italy, 2023–2025, 202232A8AN \say{Computational modeling of the heart: from eﬀicient numerical solvers to cardiac digital twins}.

The present research is part of the activities of “Dipartimento di Eccellenza 2023–2027”, MUR, Italy, Dipartimento di Matematica, Politecnico di Milano.

AT and LD acknowledge the valuable suggestions on the methodological analysis of this paper provided by professors Alfio Quarteroni and Tan Bui-Thanh.

\section*{Competing interests}
The authors declare no competing interests.

\bibliographystyle{plain}
\bibliography{references}

\end{document}